\documentclass{article} 
\usepackage{iclr2027_conference,times}

\usepackage{amsmath,amsfonts,bm}

\def\eqref#1{\textup{(\ref{#1})}}

\def\1{\bm{1}}

\DeclareMathAlphabet{\mathsfit}{\encodingdefault}{\sfdefault}{m}{sl}
\SetMathAlphabet{\mathsfit}{bold}{\encodingdefault}{\sfdefault}{bx}{n}

\newcommand{\nngs}{\ensuremath{\mathrm{NNGS}}}
\newcommand{\lid}{\ensuremath{\mathrm{LID}}}
\newcommand{\parresidue}{\ensuremath{\mathrm{PAR}}}

\newcommand{\tsi}{\ensuremath{\mathrm{TSI}}}
\newcommand{\mi}{\ensuremath{\mathrm{MI}}}

\newcommand{\HH}{\mathbf{H}}

\usepackage{hyperref}
\usepackage{url}
\usepackage{algorithm}
\usepackage{algorithmic}
\usepackage{amsmath}
\usepackage{amssymb}
\usepackage{array}
\usepackage{booktabs}
\usepackage{caption}
\usepackage{subcaption}
\usepackage{enumitem}
\usepackage{graphicx}
\usepackage{makecell}
\usepackage{natbib}
\usepackage{tabularx}
\usepackage{wrapfig}
\usepackage{xcolor}

\usepackage[most]{tcolorbox}
\tcbset{
    takeaways/.style={
        colback=yellow!12!white,
        colframe=orange!85!black,
        colbacktitle=orange!90!black,
        coltitle=white,
        boxrule=0.8pt,
        arc=4pt,
        left=4pt,
        right=4pt,
        top=2pt,
        bottom=2pt,
        fonttitle=\bfseries
    }
}

\setlist{
  labelsep=6pt,
  topsep=1pt plus 1pt minus 1pt,
  itemsep=1pt plus 1pt minus 0pt,
  parsep=0pt,
  partopsep=0pt
}

\title{Geometric Similarity in VLM Low-Level \\Vision Representations}

\newcommand{\equalcontribmark}{\textsuperscript{*}}
\newcommand{\correspondingmark}{\textsuperscript{\textdagger}}

\author{
Shao-Jun Xia \equalcontribmark \\
Duke University \\
\texttt{\small shaojun.xia@duke.edu} \\
\And
Huixin Zhang \equalcontribmark \\
Texas A\&M University \\
\texttt{\small zhanghui21@tamu.edu} \\
\And
Zhen Lei \equalcontribmark \\
Florida State University \\
\texttt{\small zl24a@fsu.edu} \\
\AND
Anlan Sun \correspondingmark \\
UC San Diego \\
\texttt{\small ans108@ucsd.edu} \\
\And
Yuner Zhang \\
University of Pennsylvania \\
\texttt{\small yunerzh@seas.upenn.edu} \\
\And
Xiaoyang Chen \correspondingmark \\
UNC Chapel Hill \\
\texttt{\small xiaoyang\_chen@med.unc.edu}
}

\iclrfinalcopy 
\begin{document}

\maketitle

{\vspace{-20pt}
\small
\textsuperscript{*}Equal contribution.\quad
\textsuperscript{\textdagger}Corresponding authors.\par
\vspace{20pt}}

\fancyhead{}
\renewcommand{\headrulewidth}{0pt}

\begin{abstract}
Vision-language models (VLMs) have emerged as powerful candidates for universal vision backbones, with representative architectures including autoregressive (AR) models and diffusion transformers (DiTs). Yet, adapting them efficiently for all-in-one low-level image restoration remains a challenge. Crucially, the field lacks an understanding of how VLMs organize hidden-layer representations and whether these structurally distinct paradigms share a common geometric organization for pixel-level perception. Such shared organization is a prerequisite for building highly transferable, unified restoration VLMs and adapters.
In this paper, we systematically investigate representational similarity across 24 low-level tasks spanning 5 categories.
We propose GeoSim, a unified four-level framework that analyzes task-conditioned representations from global similarity, local geometry, sparse feature decomposition, and topological verification perspectives.
Our formulation applies to the analysis of hidden states in AR models and feature maps in DiTs across same- and cross-task/model settings.
Our results reveal the organizing principles of low-level visual representations while exposing their limits in cross-task and cross-model agreement. Ultimately, GeoSim provides an interpretability lens for probing latent transferability in low-level vision and diagnosing model limitations in task- or model-specific scenarios.
Our code is available on \href{https://github.com/Xzcv-hub/VLM-Representation-Learning}{GitHub}.
\end{abstract}

\section{Introduction}
\label{sec:intro}

Vision-language models (VLMs) have demonstrated remarkable capabilities across a broad spectrum of visual understanding tasks~\citep{zhang2024vision}. In high-level vision, they have achieved success in image classification~\citep{radford2021learning}, object detection~\citep{li2022grounded}, semantic segmentation~\citep{li2022language}, instance segmentation~\citep{huynh2022open}, and open-vocabulary recognition~\citep{zhong2022regionclip}. More recently, VLMs have also begun to extend into low-level vision, where a single model is expected to solve diverse perceptual tasks such as restoration, removal, enhancement, reconstruction, and photometric correction~\citep{ai2024multimodal,jiang2025survey}. Unlike high-level vision---which primarily maps inputs to invariant semantic categories---low-level tasks must carefully balance fine-grained pixel statistics, frequency components, and local image structures with high-level perceptual context. Consequently, modern all-in-one restoration systems increasingly rely on shared internal representations to transfer knowledge across diverse perceptual tasks. Understanding how these representations are organized is therefore becoming a fundamental problem, with important implications for model interpretability, efficient knowledge transfer, continual learning, and the principled design of scalable all-in-one architectures.

Despite rapid progress in low-level VLMs, existing studies have focused primarily on improving task performance through prompting~\citep{potlapalli2023promptir,conde2024instructir}, agentic instruction~\citep{jiang2024autodir,chen2024restoreagent}, and multimodal representations~\citep{luo2024controlling}. Comparatively little is known about the internal organization of the representations that enable these models to solve diverse perceptual tasks. In particular, it remains unclear whether different low-level tasks rely on shared computational mechanisms, whether they naturally organize into coherent perceptual families, or whether each task occupies an independent region of representation space. Without understanding this organization, it remains difficult to explain why knowledge transfers across some tasks but not others, how parameter sharing should be designed, or whether a single representation space can effectively support increasingly diverse low-level capabilities.

The geometry of hidden representations provides a powerful lens for studying the internal organization of deep models. Classical representation analysis methods, including RSA~\citep{kriegeskorte2008representational}, SVCCA~\citep{raghu2017svcca}, and CKA~\citep{kornblith2019similarity}, characterize neural representations through pairwise similarity, shared linear subspaces, or relational structure. In VLMs, contrastive models such as CLIP and ALIGN learn a shared embedding space by encouraging matched image--text pairs to exhibit higher similarity than mismatched pairs~\citep{radford2021learning,jia2021scaling}. Subsequent work has shown that large multimodal models develop increasingly shared representational structures across architectures and, in some cases, across modalities~\citep{huh2024platonic,jha2026harnessing}.
More recent layer-wise analyses further reveal that visual representations become progressively aligned with textual feature spaces~\citep{venhoff2025visual}, object-specific information is refined through subsets of visual tokens~\citep{neo2025towards}, and visual information is increasingly transferred into text-token representations in deeper layers~\citep{lin2025boosting}.

Complementary advances in interpretability suggest that concepts, decisions, and intermediate computations are often organized as linear directions or low-dimensional manifolds in hidden space~\citep{marks2023geometry,joshi2026geometry,gurnee2026models,ma2026reasoning}. Sparse autoencoders (SAEs) further decompose hidden activations into interpretable sparse features~\citep{huben2024sparse,gao2025scaling,pach2026sparse,shu2025survey}, while recent evidence indicates that semantically equivalent visual and textual inputs can induce shared modality-invariant task representations~\citep{luo2025vision}. Collectively, these studies suggest that neural representations often possess rich geometric structure. However, nearly all existing analyses focus on semantic concepts, language alignment, or simplified multimodal architectures.
Whether similar geometric principles govern perceptual representations in low-level vision, and how they extend from object and modality comparisons to task-conditioned low-level transformations, remains largely unexplored.
Understanding whether such tasks exhibit a coherent geometric organization is therefore important not only for interpretability but also as a foundation for predicting task transfer, designing modular architectures, identifying shared computational primitives, and scaling unified restoration models to increasingly diverse perceptual capabilities.

Motivated by these observations, this paper investigates the geometric organization of low-level visual task representations in modern VLMs. Specifically, we ask:

\begin{itemize}
    \item[\textbf{RQ1.}]
    Do low-level visual tasks exhibit meaningful geometric clustering in VLM hidden representations, and how does this organization evolve across network depth?

    \item[\textbf{RQ2.}]
    Do low-level task representations exhibit shared and interpretable latent structures within and across task families?

    \item[\textbf{RQ3.}]
    To what extent is the geometric organization of low-level tasks universal across different VLM architectures?
\end{itemize}

\begin{figure*}[t]
    \centering
    \includegraphics[width=\textwidth]{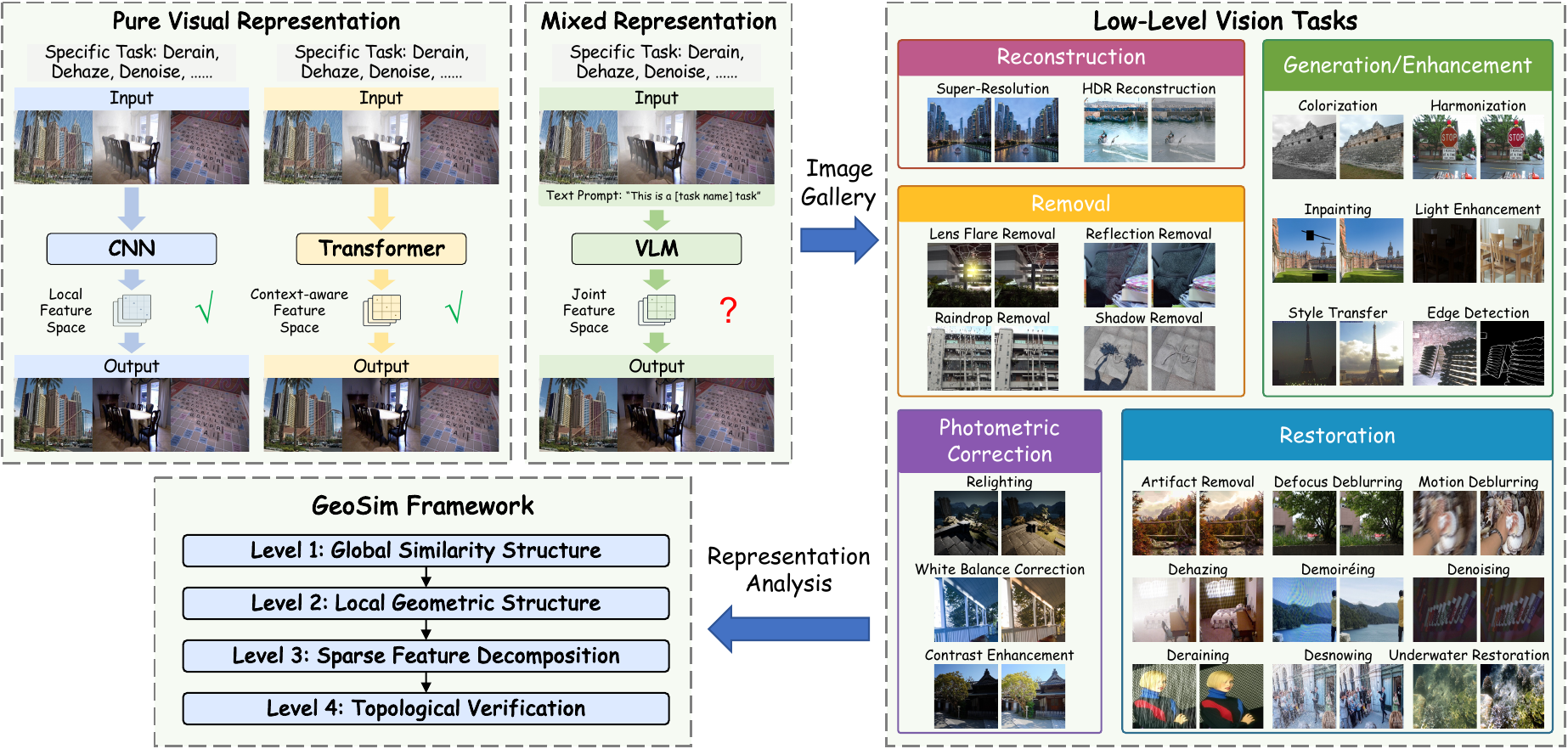}
    \caption{Overview of VLM representations for low-level tasks and the GeoSim framework.}
    \vspace{-10pt}
    \label{fig:figure1}
\end{figure*}

To answer these questions, we introduce \textbf{GeoSim} (Figure~\ref{fig:figure1}), a unified four-level framework for characterizing the geometry of low-level visual task representations.
GeoSim analyzes task-conditioned representations from four complementary perspectives: global similarity structure, local geometric structure, sparse feature decomposition, and topological verification.
This unified framework enables systematic comparisons across tasks, layers, architectures, and model families, providing a comprehensive view of how perceptual representations are organized within modern VLMs.

Our main contributions are summarized as follows:

\begin{itemize}[leftmargin=11.2pt]
    \item We present a systematic geometric study of \textbf{24 low-level vision tasks across multiple VLM families} and establish a reusable benchmark with unified prompt templates and evaluation protocols.
    \item We reveal how task geometry evolves across depth and quantify global, local, and cross-model agreement with Inter-task/Intra-task Cosine Similarity, Distance Correlation ($\mathrm{dCor}$), Nearest Neighbor Graph Similarity ($\mathrm{NNGS}$), Local Intrinsic Dimensionality ($\mathrm{LID}$), Procrustes Alignment Residual ($\mathrm{PAR}$), and topological analyses.
    \item We establish a comprehensive pipeline including Sparse Autoencoder ($\mathrm{SAE}$)-based feature decomposition, Mutual Information ($\mathrm{MI}$), and cross-model Maximum Mean Discrepancy ($\mathrm{MMD}$) to distinguish task-specific and cross-task shared sparse representations.
\end{itemize}

\section{Related Work}
\label{sec:related}

\paragraph{All-in-one image restoration.}
Image restoration has gradually evolved from task-specific networks toward unified models capable of handling diverse degradations within a single architecture~\citep{jiang2025survey}.
Recent work leverages VLM representations~\citep{luo2024controlling}, visual instructions with diffusion priors~\citep{luo2025visual}, and autoregressive visual token generation~\citep{rajagopalan2025restorevar} for restoration.
These advances demonstrate that increasingly diverse restoration tasks can be handled within a shared representation space, yet the internal organization of the learned representations remains largely unexplored.
Beyond restoration, \citet{li2025investigate} used component-wise tuning to show that distortion recognition in an MLLM depends most strongly on adapting its visual encoder, with further gains from the alignment projector. This component-level evidence motivates our study of how low-level tasks are organized across layers and models.

\vspace{-8pt}
\paragraph{Task transferability and task relationships.}
Understanding relationships among visual tasks has long been a central topic in computer vision because tasks with similar representations are expected to transfer knowledge more effectively.
Taskonomy~\citep{zamir2018taskonomy} systematically quantified transfer relationships among a large collection of visual tasks, Task2Vec~\citep{achille2019task2vec} proposed compact task embeddings for directly measuring task similarity, and later studies leveraged representation similarity to construct task taxonomies and guide multi-task learning~\citep{dwivedi2019crosstask,standley2020mtl}.
\citet{sachdeva2026understanding} provided a comprehensive study of positive and negative transfer across visual perception tasks in VLMs.
Beyond explicit transfer, recent work demonstrated that VLMs can spontaneously develop cross-modal task representations from in-context examples~\citep{luo2025vision} and perform cross-task visual in-context learning through implicit text-driven prompts~\citep{xia2025t2t}.
While these studies characterize \emph{whether} tasks transfer knowledge, they provide limited insight into \emph{why} such transfer occurs. In particular, how transferability is reflected in the geometric organization of hidden representations within modern VLMs remains largely unknown.

\vspace{-8pt}
\paragraph{Geometric representation analysis.}
Geometric representation analysis seeks to understand neural networks by characterizing the structure of their hidden representations.
These techniques have recently been applied to VLMs to study shared image--text embedding spaces~\citep{radford2021learning}, cross-model representational convergence~\citep{huh2024platonic}, language-space interpretation of visual features~\citep{jiang2025interpreting}, modality-invariant task representations~\citep{luo2025vision}, and progressive visual--language alignment~\citep{venhoff2025visual}. Nevertheless, most existing studies focus on semantic concepts, language alignment, or individual geometric properties.
A unified framework that systematically integrates global, local, distributional, sparse, and topological analyses for understanding \emph{low-level perceptual task representations} remains lacking.

\vspace{-8pt}
\paragraph{SAEs for model interpretability.}
SAEs have emerged as an effective tool for interpreting hidden representations in large neural networks~\citep{huben2024sparse,gao2025scaling,templeton2026scaling}.
Recently, SAE-based analysis has been applied to vision and vision-language models.
Universal SAEs learn a single sparse concept space jointly across several vision encoders~\citep{thasarathan2025universal}, whereas Matryoshka SAEs optimize nested sparsity levels within CLIP representations~\citep{zaigrajew2025interpreting}.
\citet{pach2026sparse} further demonstrated that SAEs can recover monosemantic and steerable features from VLM representations.
These studies establish SAEs as an effective tool for interpreting individual concepts within VLMs. While in our work, we employ SAEs to characterize shared and task-specific sparse representations across multiple low-level vision tasks and multiple VLM architectures, providing a feature-level perspective on the geometric organization of perceptual task representations.
Extended related work is provided in Appendix~\ref{app:related}.

\section{Proposed Framework}
\label{sec:framework}

\subsection{Framework Overview}

We introduce \textbf{GeoSim}, a unified framework for characterizing the geometric organization of low-level visual task representations in VLMs. Rather than relying on a single similarity measure, GeoSim analyzes representations from four complementary perspectives:
Level~1 characterizes the global similarity structure among tasks and its evolution across depth, Level~2 examines local geometric structure, Level~3 decomposes representations into shared and task-specific sparse features, and Level~4 examines cross-model universality.
Notation is summarized in Table~\ref{tab:notation}, and the motivation for each level is detailed in Appendix~\ref{app:level-roles}.

\begin{table}[t]
\centering
\caption{Notation used throughout Section~\ref{sec:framework}.}
\label{tab:notation}
\small
\renewcommand{\arraystretch}{1.2}
\begin{tabularx}{\linewidth}{@{} l >{\raggedright\arraybackslash}X @{}}
\toprule
\textbf{Symbol} & \textbf{Meaning} \\
\midrule
$x$ & Input image \\
$n, b$ & Samples per task in each sampling repeat ($100$); number of sampling repeats ($10$) \\
$i, j, s$ & Image indices ($i,j$); sampling-repeat index ($s$) \\
$p, q, t \in \mathcal{T}$ & Task indices ($p,q$ for pairs, $t$ for single task) \\
$\ell, d, d_{\mathrm{PCA}}$ & Layer index ($\ell^*$: stable layer); feature dimension ($d$); retained principal-component dimension ($d_{\mathrm{PCA}}$) \\
$h^\ell_p(x) \in \mathbb{R}^d$ & Representation of image $x$ for task $p$ at layer $\ell$ \\
$\mathbf{H}^\ell_p \in \mathbb{R}^{n\times d}$ & Task representation matrix $[h^\ell_p(x_1); \dots; h^\ell_p(x_n)]$ \\
$M_1, M_2$ & Evaluated VLMs \\
$f$ & SAE feature index\\
$\mathcal{O}(d_{\mathrm{PCA}})$ & Group of $d_{\mathrm{PCA}}\times d_{\mathrm{PCA}}$ orthogonal matrices \\
$\|\cdot\|_F$ & Frobenius norm \\
\bottomrule
\end{tabularx}
\vspace{-12pt}
\end{table}

\subsection{Level 1: Global Similarity Structure}
\label{sec:level1}

\paragraph{Inter-Task and Intra-Task similarity.}
For an input image $x$, let $h_p^\ell(x)\in\mathbb{R}^{d}$ denote the hidden representation extracted for task $p$ at layer $\ell$.
Given $n$ sampled images, we stack their representations into $\HH_p^\ell=[h_p^\ell(x_1);\ldots;h_p^\ell(x_n)]\in\mathbb{R}^{n\times d}$.
$\HH_p^{\ell}$ is formed in each sampling repeat $s$, with $s$ suppressed when unambiguous.
We compute the average inter-task cosine similarity between every pair of tasks
\begin{equation}
S^{\ell}(p,q)
=
\frac{1}{b}
\sum_{s=1}^{b}
\frac{1}{n^2}
\sum_{i=1}^{n}
\sum_{j=1}^{n}
\cos
\left(
h_p^\ell(x_i^s),
h_q^\ell(x_j^s)
\right),
\label{eq:inter-task-pair-cosine}
\end{equation}
where $b$ denotes the number of sampling repeats. Given a single task $t$, we sample $n$ distinct images. The average intra-task cosine similarity is defined as
\begin{equation}
S^{\ell}(t)
=
\frac{1}{b}
\sum_{s=1}^{b}
\frac{1}{n^2}
\sum_{i=1}^{n}
\sum_{j=1}^{n}
\cos
\left(
h_t^\ell(x_i^s),
h_t^\ell(x_j^s)
\right).
\label{eq:intra-task-pair-cosine}
\end{equation}

Collecting all pairwise similarities produces layer-wise task similarity matrix $\mathbf{S}^{\ell}=[S^{\ell}(p,q)]_{p,q\in\mathcal{T}}$, which summarizes the global organization of low-level task representations at layer $\ell$.

\vspace{-8pt}
\paragraph{Cross-model similarity.}
Cosine similarity characterizes relationships between tasks within a single model but does not directly compare the geometric structure of representations across architectures. We therefore additionally employ dCor~\citep{szekely2007measuring}, which measures statistical dependence between two representation matrices based on pairwise distance structure.

For the same task $t$, let $\mathbf H_t^{\ell,M_1}$ and $\mathbf H_t^{\ell,M_2}$ denote the representation matrices extracted from models $M_1$ and $M_2$, respectively. In cross-model expressions, $\ell$ refers to the selected layer $\ell^*_{M}$ of each model, which need not coincide across models.

The overall cross-model statistic is defined as the average dCor over the $b$ sampling repeats:
\begin{equation}
\mathrm{dCor}^{\ell}(M_1,M_2)
=
\frac{1}{b}
\sum_{s=1}^{b}
\frac{
\mathrm{dCov}^{s}
}{
\sqrt{
\mathrm{dVar}_1^{s}\,
\mathrm{dVar}_2^{s}
}
}.
\label{eq:dcor}
\end{equation}
Here, $\mathrm{dCov}^{s}$ and $\mathrm{dVar}_1^{s},\mathrm{dVar}_2^{s}$ denote the distance covariance and distance variances in repeat $s$, computed from double-centered pairwise Euclidean distances as detailed in Appendix~\ref{app:dcor-details}.

\vspace{-8pt}
\paragraph{Layer-wise evolution.}
Representation geometry evolves substantially across transformer layers. To identify where the global organization of low-level tasks stabilizes, we compute the Frobenius distance $D_{\mathrm{full}}^{\ell}=\|\mathbf{S}^{\ell}-\mathbf{S}^{\ell+1}\|_F$ between consecutive task similarity matrices.
A large value of $D_{\mathrm{full}}^{\ell}$ indicates substantial reorganization of task relationships between adjacent layers, whereas a consistently small value indicates a stable task similarity structure. We designate the final layer within this stable region as $\ell^*$ and use it for the subsequent analyses (Appendix~\ref{app:extraction}).

\vspace{-8pt}
\paragraph{Visualization.}
Node positions in the task relation graph (Figure~\ref{fig:figure5}) are obtained by classical multidimensional scaling (MDS)~\citep{torgerson1952multidimensional} of the cross-model consensus dissimilarities derived from $S^{\ell}(p,q)$.

\subsection{Level 2: Local Geometric Structure}
\label{sec:level2}

\paragraph{Local intrinsic dimensionality (\lid).}
We first characterize the local complexity of each task representation space using \lid, estimated by maximum likelihood~\citep{levina2004maximum} from the $k$ nearest neighbors of each sample within the same task and averaged over samples and sampling repeats (Appendix~\ref{app:lid}).

\vspace{-8pt}
\paragraph{$k$-nearest-neighbor graph similarity (\nngs).}
To directly evaluate local neighborhood preservation across models, we construct $k$-nearest-neighbor graphs over corresponding samples and measure the fraction of shared neighbors. For representation matrices $U$ and $V$ ($\HH_t^{\ell, M_1},\HH_t^{\ell, M_2}$) extracted from two models, the neighborhood similarity~\citep{huh2024platonic} is
\begin{equation}
\nngs(U,V)
=
\frac{1}{b}
\sum_{s=1}^{b}
\frac{1}{n}
\sum_{i=1}^{n}
\frac{
|N_k(x_i^s;U)\cap N_k(x_i^s;V)|
}{k},
\label{eq:nngs}
\end{equation}
where $N_k(x_i^s;U)$ denotes the $k$ nearest neighbors of sample $i$ in representation space $U$.

\vspace{-8pt}
\paragraph{Procrustes alignment residual (\parresidue).}
Neighborhood similarity evaluates local topology, but does not capture whether two representation spaces can be globally aligned after removing coordinate differences. Therefore, we further measure geometric alignment using orthogonal Procrustes analysis~\citep{schonemann1966generalized}.

For task $t$ and models $M_1,M_2$, we first center each representation matrix and apply principal component analysis (PCA) separately, retaining $d_{\mathrm{PCA}}\leq\min(d_{M_1},d_{M_2},n-1)$ dimensions. The resulting matrices are denoted as $\HH_t^{\ell, M_1},\HH_t^{\ell, M_2}\in\mathbb{R}^{n\times d_{\mathrm{PCA}}}$. The alignment residual is defined as
\begin{equation}
\parresidue(t,M_1,M_2)
=
\min_{Q\in\mathcal{O}(d_{\mathrm{PCA}})}
\|
\HH_t^{\ell, M_1}
-
\HH_t^{\ell, M_2}Q
\|_F .
\label{eq:procrustes}
\end{equation}
A smaller $\parresidue$ indicates stronger geometric agreement up to an orthogonal transformation.

\subsection{Level 3: Sparse Feature Decomposition}
\label{sec:level3}

\paragraph{SAE representation decomposition.}
At the stable layer $\ell^*$, we write $z_t(x)=h_t^{\ell^*}(x)\in\mathbb{R}^{d}$.
For each VLM, we train a single ReLU SAE with an $\ell_1$ sparsity penalty on $z_t(x)$, shared across all low-level vision tasks, so that feature usage can be compared within a common learned dictionary;
$a(x)\in\mathbb{R}^{D_{\mathrm{SAE}}}$ denotes its sparse code and $D_{\mathrm{SAE}}$ the dictionary size (Appendix~\ref{app:level3-details}; training details in Appendix~\ref{app:sae}).
The SAE is trained on task-balanced samples, and all feature statistics are computed on held-out samples.

\vspace{-8pt}
\paragraph{Feature activation statistics.}
For SAE feature $f$, we define its activation rate for task $t$ as
\begin{equation}
\rho_f(t)
=
\frac{1}{N_t}
\sum_{x\in\mathcal{D}_t}
\mathbf{1}
[a_f(x)>\tau],
\label{eq:feature-rate}
\end{equation}
where $\mathcal D_t$ is the held-out sample set for task $t$, $N_t=|\mathcal D_t|$, $a_f(x)$ denotes the activation value of feature $f$, and $\tau$ is an activation threshold. The activation rate $\rho_f(t)$ measures how frequently a feature is used by a given task.

\vspace{-8pt}
\paragraph{Task selectivity index (\tsi).}
We quantify whether a feature is concentrated on a small number of tasks using \tsi:
\begin{equation}
\tsi(f)=\frac{\max_{t\in\mathcal{T}}\rho_f(t)}{\sum_{t\in\mathcal{T}}\rho_f(t)}.
\label{eq:tsi}
\end{equation}
A high $\tsi(f)$ indicates that feature $f$ is preferentially activated by a particular task, whereas a lower value indicates broader usage across tasks. Here, $\rho_f(t)$ is the activation probability defined in Eq.~\eqref{eq:feature-rate}. Features are categorized as task-specific or shared according to predefined thresholds $\theta_{\mathrm{high}}$ and $\theta_{\mathrm{low}}$.

\vspace{-8pt}
\paragraph{Feature--task mutual information (\mi).}
To quantify whether individual SAE features encode task identity, we measure the \mi{} between feature activation and task labels. For each SAE feature $f$, we define its binary activation state as
$Z_f=\mathbf{1}[a_f(x)>\tau]$,
where $\tau$ is the activation threshold, $Z_f\in\{0,1\}$. The \mi{} between feature activation and task identity is computed as
\begin{equation}
I(Z_f;T)
=
\sum_{z\in\{0,1\}}
\sum_{t\in\mathcal{T}}
P(z,t)
\log
\frac{P(z,t)}{P(z)P(t)},
\label{eq:mi-feature-task}
\end{equation}
where $P(z,t)$ is estimated from held-out samples across all tasks. Higher values indicate that a feature activation pattern carries more information about task identity, suggesting stronger task selectivity.

\vspace{-8pt}
\paragraph{Cross-model sparse representation comparison.}
To evaluate whether different VLMs develop similar sparse representations, we compare feature activation-rate distributions across models. For a given task $t$, let $A_t^{M_1}$ and $A_t^{M_2}$ denote the empirical distributions of the activation rates $\rho_f(t)$ obtained from the SAEs of models $M_1$ and $M_2$, respectively. Each feature active on at least one held-out task contributes one scalar rate, while features inactive across all held-out tasks are excluded. We measure their distributional similarity using the MMD~\citep{gretton2012kernel} with a radial basis function (RBF) kernel $\kappa$:
\begin{equation}
\operatorname{MMD}^2
(A_t^{M_1},A_t^{M_2})
=
\left\|
\mathbb E_{A_t^{M_1}}\phi
-
\mathbb E_{A_t^{M_2}}\phi
\right\|_{\mathcal H_\kappa}^{2},
\label{eq:mmd}
\end{equation}
where $\phi$ denotes the feature mapping induced by kernel $\kappa$, $\mathbb{E}_{A_t^{M_1}}\phi$, $\mathbb{E}_{A_t^{M_2}}\phi$ are shorthand for
$\mathbb{E}_{\rho\sim A_t^{M_1}}[\phi(\rho)]$, $\mathbb{E}_{\rho\sim A_t^{M_2}}[\phi(\rho)]$,
with $\rho\in[0,1]$ denoting a feature activation rate. We report the squared MMD between these empirical distributions. Lower values indicate more similar task-conditioned activation-rate distributions, without establishing correspondence between individual SAE features across models.

\subsection{Level 4: Topological Verification}
\label{sec:level4}

For each task $t$, we project the stable-layer representations $\HH_t^{\ell^*}$ to a low-dimensional point cloud $\widetilde{\HH}_t\subset\mathbb{R}^{d_{\mathrm{PH}}}$ with a fixed UMAP~\citep{mcinnes2018umap} configuration ($d_{\mathrm{PH}}\ll d$), which mitigates distance concentration in the ambient space.
Using persistent homology (PH), we summarize the birth and death of connected components and one-dimensional loops (homology degrees $H_0$ and $H_1$) in persistence diagrams, and compare models on the same task with the 2-Wasserstein and bottleneck distances between their diagrams.
Level~4 serves as a supplementary cross-check; construction details and results are given in Appendices~\ref{app:topology-details} and~\ref{app:vis}.

\section{Experiments}

\subsection{Experimental Setup}
\label{sec:setup}

\paragraph{Datasets.}
We construct a benchmark suite containing 24 low-level vision tasks grouped into five perceptual families: restoration, removal, generation/enhancement, reconstruction, and photometric correction (see Table~\ref{tab:tasks}).
For tasks such as artifact removal, colorization, and inpainting, we construct datasets by synthesizing degraded inputs from clean images.
Dataset sources and construction details are provided in Appendix~\ref{app:data_details}.

For each task, we sample $n=100$ distinct input images from representative low-level vision datasets in each of $b=10$ fixed sampling repeats. To minimize the influence of instruction variation, we use a fixed prompt template for each task and keep the prompt format consistent across models.

\begin{table}[t]
\centering
\caption{Low-level vision task suite organized by task family.}
\label{tab:tasks}
\scriptsize
\setlength{\tabcolsep}{3pt}
\renewcommand{\arraystretch}{1.05}
\begin{tabularx}{\linewidth}{
@{}
>{\raggedright\arraybackslash}p{2.8cm}
>{\raggedright\arraybackslash}X
@{}
}
\toprule
\textbf{Family} & \textbf{Tasks} \\
\midrule
Restoration &
Artifact Removal, Defocus Deblurring, Motion Deblurring, Dehazing, Demoir\'eing, Denoising, Deraining, Desnowing, Underwater Restoration \\
Removal &
Lens Flare Removal, Raindrop Removal, Reflection Removal, Shadow Removal \\
{Generation/Enhancement} &
Colorization, Harmonization, Inpainting, Light Enhancement, Style Transfer, Edge Detection \\
Reconstruction &
Super-Resolution, HDR Reconstruction \\
Photometric Correction &
Relighting, White Balance Correction, Contrast Enhancement \\
\bottomrule
\end{tabularx}
\vspace{-8pt}
\end{table}

\vspace{-8pt}
\paragraph{Implementation Details.}
We analyze unified VLMs spanning visual understanding and image generation across multiple architectural families and parameter scales (see Table~\ref{tab:models}).
All experiments were conducted using eight NVIDIA RTX 6000 GPUs, each with 48 GB of memory, and eight NVIDIA B200 GPUs, each with 180 GB of memory.

\subsection{Results}
\label{sec:result}

Our main results are organized according to the four task/model conditions, with the metrics in Appendix Table~\ref{tab:map}.

\subsubsection{Condition I: Same Task, Same Model}

\begin{figure}[t]
    \centering
    \includegraphics[width=\columnwidth]{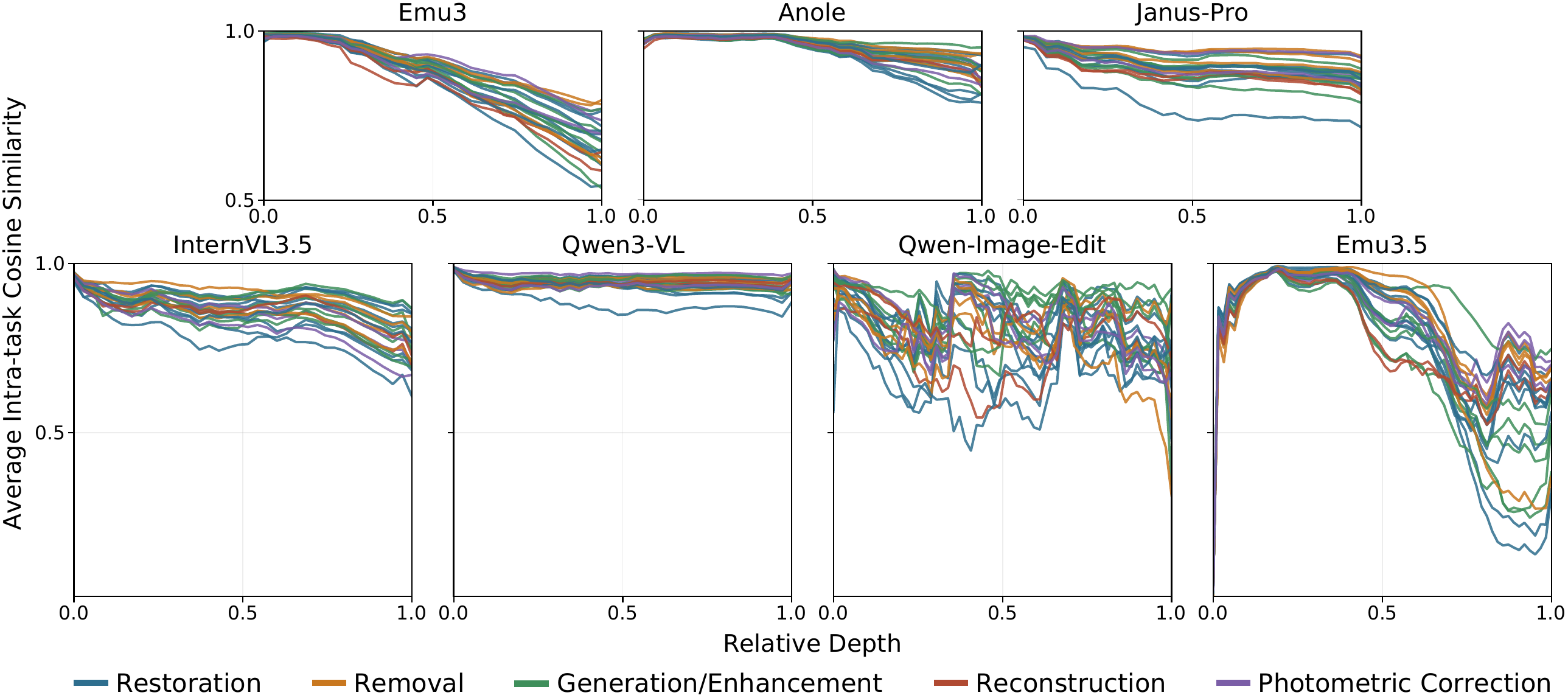}
    \caption{\textbf{Condition I:} Consistency of the instruction-induced displacement across network depth. Each panel is one model; each of the 24 curves is one task, colored by perceptual family.}
    \label{fig:figure2}
    \vspace{-10pt}
\end{figure}

\begin{wraptable}{r}{0.42\columnwidth}
\vspace{-40pt}
\centering
\caption{VLMs used in our analyses.}
\label{tab:models}
\scriptsize
\setlength{\tabcolsep}{2pt}
\begin{tabular}{l p{4.4cm}}
\toprule
\textbf{Scale} & \textbf{Models} \\
\midrule
\makecell[lt]{Base \\($\sim$7--8B)} &
\makecell[lt]{Emu3-Chat (8B)~\citep{wang2024emu3,wang2026emu3}, \\Anole-7B~\citep{chern2024anole}, \\Janus-Pro-7B~\citep{chen2025januspro}, \\InternVL3.5-8B~\citep{wang2025internvl35}, \\Qwen3-VL-8B~\citep{bai2025qwen3vl}} \\
\midrule
\makecell[lt]{Large \\($>$13B)} &
\makecell[lt]{Qwen-Image-Edit ($\sim$20B)~\citep{wu2025qwenimage}, \\Emu3.5-Image ($\sim$34B)~\citep{cui2025emu35}} \\
\bottomrule
\end{tabular}
\vspace{-15pt}
\end{wraptable}
Condition I fixes the task and model while varying the input image. As shown in Figure~\ref{fig:figure2}, all seven models exhibit aligned instruction-induced displacements, with mean cosine similarities of $0.80$--$0.96$ at the selected analysis layer.
Despite architectural differences, these transformations are stable across images at that layer, while their consistency can vary across depth.

Appendix Figure~\ref{fig:figure1_Appendix} shows that local geometric complexity is task-dependent. The 24 tasks follow a consistent ordering across models ($\rho=0.80$), with Denoising having the lowest LID in every model. Each task exhibits a stable global direction and a characteristic local manifold complexity.

\vspace{-3pt}
\begin{tcolorbox}[
    takeaways,
    title=Key Takeaway I,
    colbacktitle=cyan!60!black
]
Low-level visual tasks exhibit stable within-task geometry, with aligned instruction-induced transformations across images and consistently organized local task manifolds across models.
\end{tcolorbox}
\vspace{-5pt}

\subsubsection{Condition II: Same Task, Different Models}

Condition II compares the same task across different models to assess cross-model geometric consistency. As shown in Figure~\ref{fig:figure3}, the 21 model pairs exhibit strong global agreement (mean $\mathrm{dCor}=0.77$), but weaker local consistency (mean chance-adjusted $\mathrm{NNGS}=0.27$). 
The held-out, normalized PAR exceeds the mean-prediction baseline of $1.0$ for 16 of 21 pairs. Overall, models share global relational geometry but not local neighborhoods or a common coordinate frame.

\begin{figure}[t]
    \centering
    \includegraphics[width=\columnwidth]{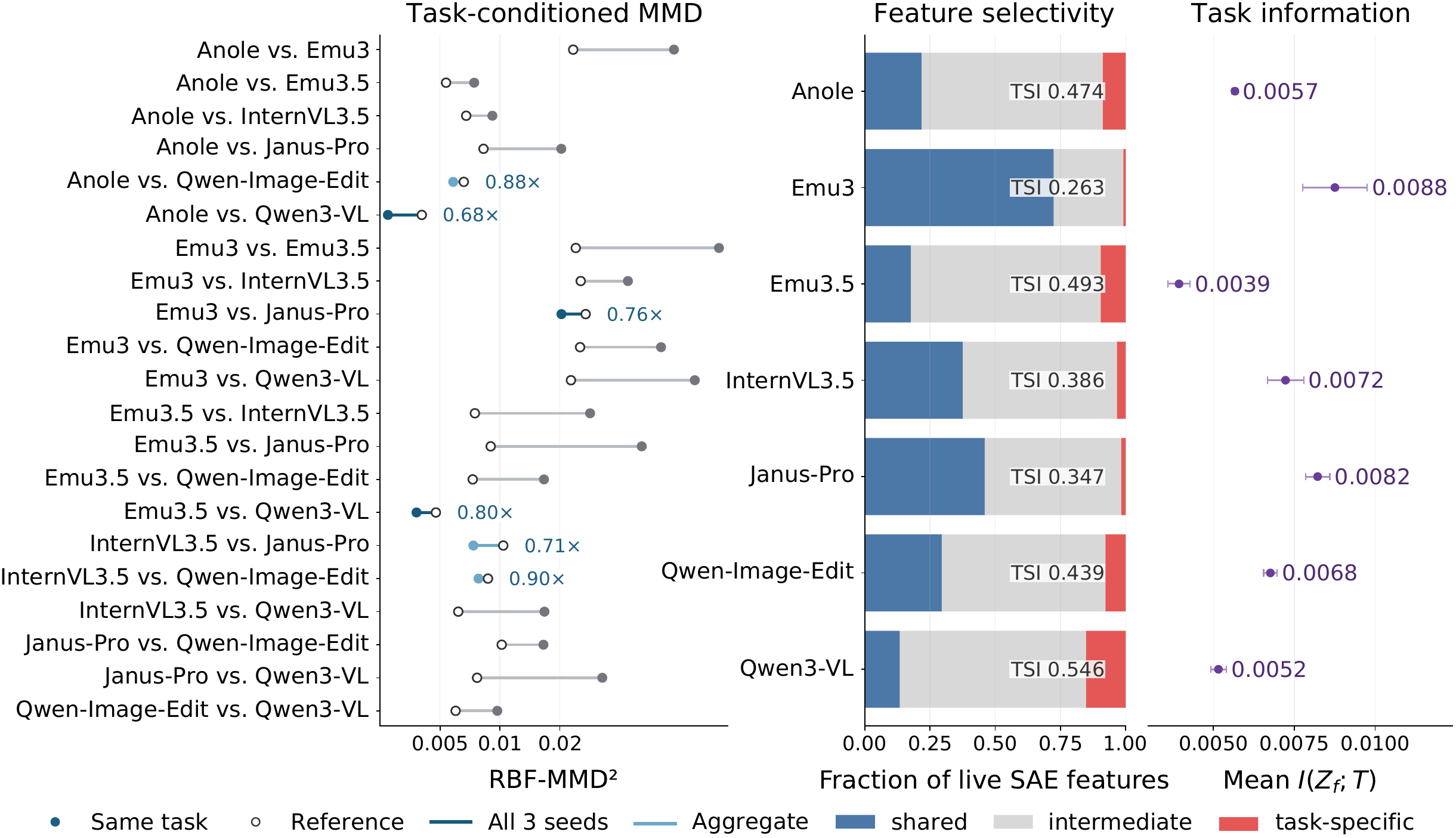}
    \caption{\textbf{Condition II:}
    Task-conditioned cross-model and within-model reference MMD (left), per-model TSI composition (center), and feature--task MI (right).}
    \label{fig:figure4}
    \vspace{-5pt}
\end{figure}

Level-3 SAE evidence in Figure~\ref{fig:figure4} further narrows the scope of this cross-model agreement by demonstrating that same-task cross-model activation-rate MMD is lower than the matched within-model, different-task reference for only 6 of 21 model pairs when aggregated over 24 tasks and three SAE seeds, with only Anole--Qwen3-VL, Emu3--Janus-Pro, and Emu3.5--Qwen3-VL satisfying this relation across all three seeds.
Moreover, model pairs with high global agreement do not generally show lower same-task MMD than the different-task reference.
Across models, task-specific features account for only 0.9--15.1\% of live features, while the shared-feature fraction varies from 13.4\% to 72.4\%, indicating that sparse features are broadly reused across tasks within each model.

\vspace{-3pt}
\begin{tcolorbox}[
    takeaways,
    title=Key Takeaway II,
    colbacktitle=cyan!60!black
]
Cross-model universality is primarily global in that models preserve within-task relational geometry despite limited local-neighborhood agreement and the absence of a shared coordinate frame. Level-3 SAE evidence shows the same boundary at the activation-rate distribution level, where 6 of 21 model pairs exhibit lower same-task MMD than the different-task reference.
\end{tcolorbox}
\vspace{-5pt}

\subsubsection{Condition III: Different Tasks, Same Model}

\begin{figure}[t]
    \centering
    \begin{minipage}[t]{0.3\columnwidth}
        \centering
        \includegraphics[width=\textwidth]{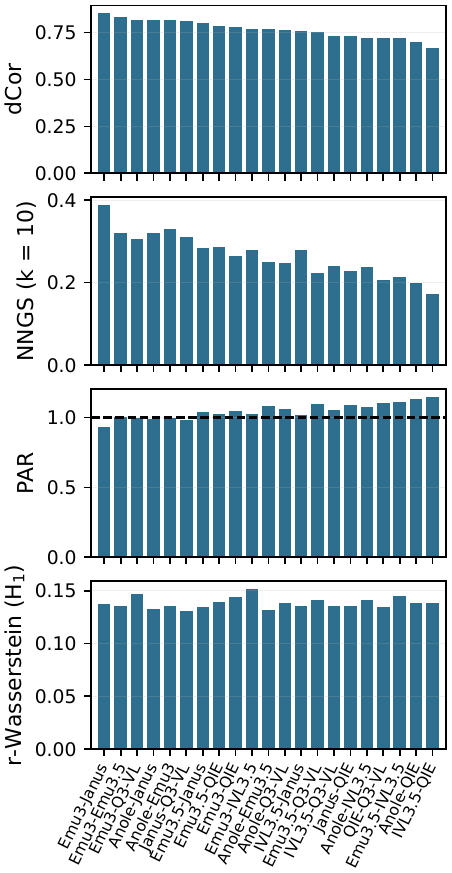}
        \captionsetup{width=\textwidth}
        \caption{\textbf{Condition II:} All 21 model pairs, averaged over 24 tasks and ordered by dCor.}
        \label{fig:figure3}
    \end{minipage}%
    \hfill
    \begin{minipage}[t]{0.698\columnwidth}
        \centering
        \includegraphics[width=\textwidth]{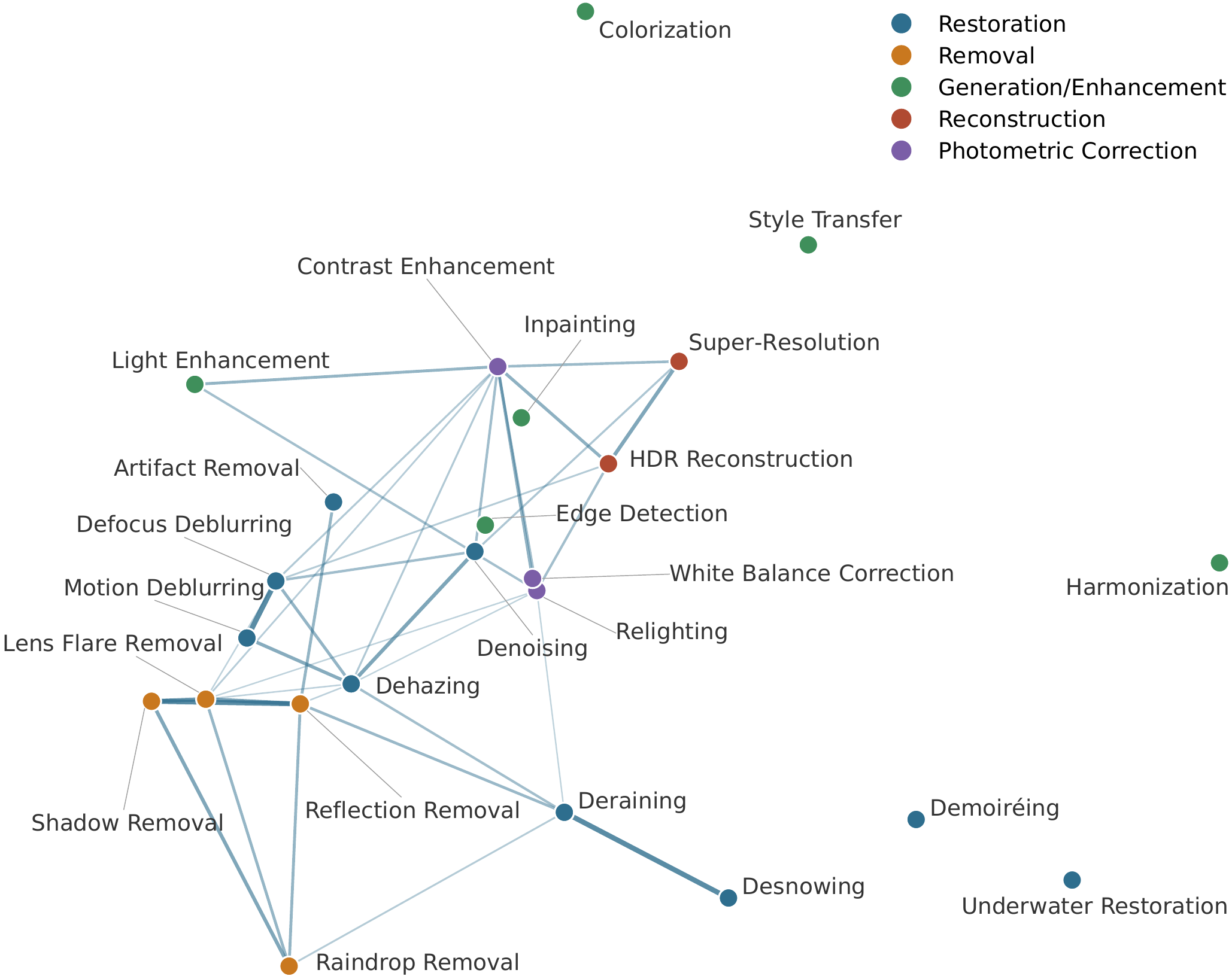}
        \captionsetup{width=0.8\textwidth}
        \caption{\textbf{Condition III:} Consensus relation graph over the 24 tasks. Node positions are visualized with MDS; edge width increases with consensus similarity.}
        \label{fig:figure5}
    \end{minipage}
\vspace{-8pt}
\end{figure}

Condition III examines how low-level visual tasks are organized within a shared model representation. The consensus graph in Figure~\ref{fig:figure5} identifies 36 reliable task relations, including 20 cross-family links. Removal and Photometric Correction tasks form the most densely connected region, with Contrast Enhancement the most connected task, while Reconstruction lies nearby and the nine Restoration tasks spread more thinly across the graph. In contrast, most Generation/Enhancement tasks (e.g., Colorization, Harmonization, and Style Transfer) are more isolated from the main cluster.

The heatmaps in Figure~\ref{fig:figure3_Appendix} (Appendix) show reproducible task structure across models, but the within- versus across-family gap is only $0.135$ standard deviations (SD), below the typical cross-model disagreement of $0.71$-SD. Thus, task geometry is better explained by shared visual mechanisms than by broad perceptual families.

\vspace{-3pt}
\begin{tcolorbox}[
    takeaways,
    title=Key Takeaway III,
    colbacktitle=cyan!60!black
]
Cross-task geometry is meaningful but mechanism-centric, with tasks sharing degradation or restoration operations forming the strongest neighborhoods, whereas the predefined perceptual families only weakly capture the representation structure.
\end{tcolorbox}
\vspace{-5pt}

\subsubsection{Condition IV: Different Tasks, Different Models}

\begin{figure}[t]
    \centering
    \includegraphics[width=\columnwidth]{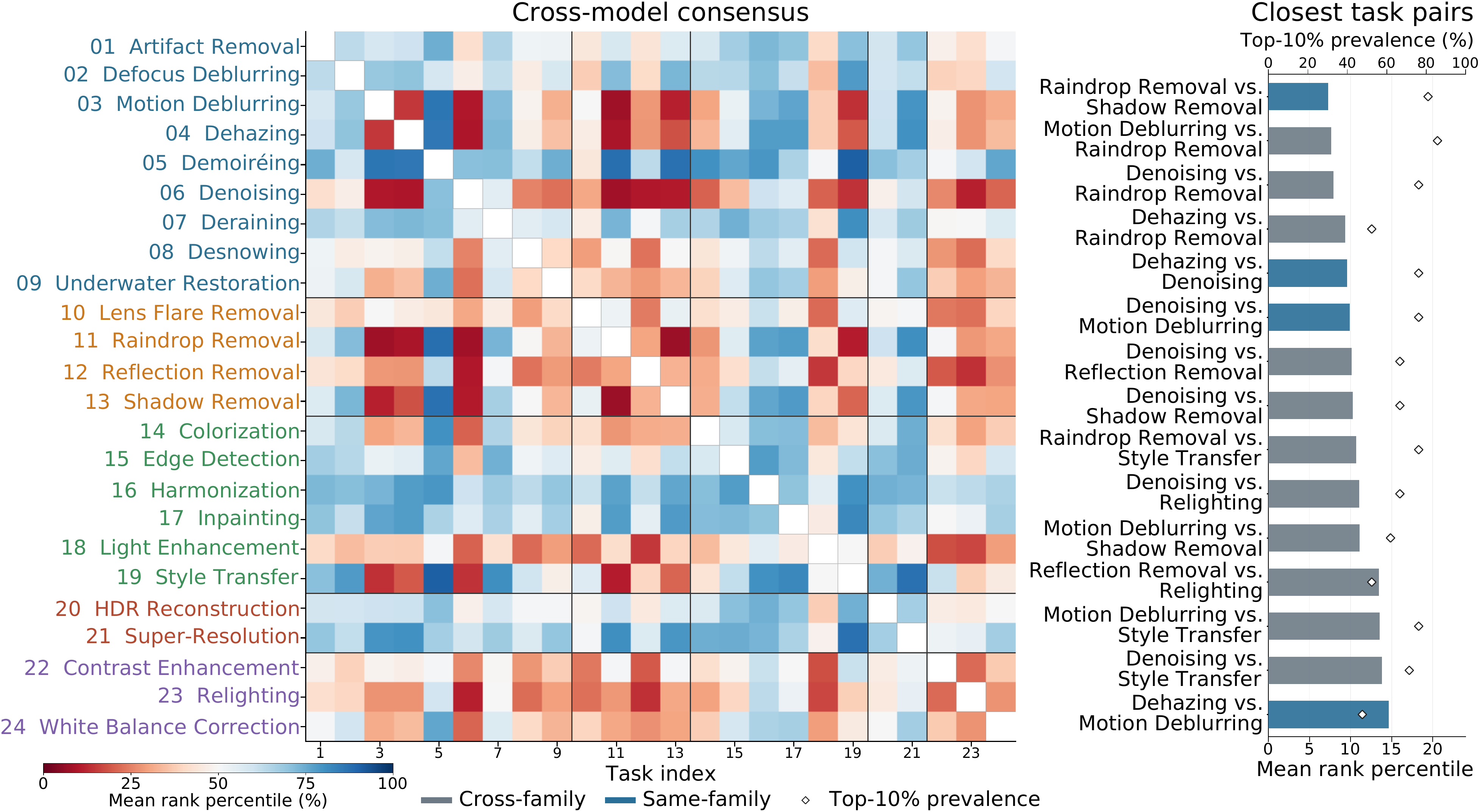}
    \caption{\textbf{Condition IV:} Cross-task, cross-model activation-rate distribution similarity from the Level-3 sweep. Results are aggregated using MMD across seven models and 24 tasks.
    }
    \label{fig:figure6}
    \vspace{-10pt}
\end{figure}

Condition IV explores whether feature activation-rate distributions are similar across tasks and models through an exhaustive Level-3 MMD sweep. In Figure~\ref{fig:figure6}, the left panel summarizes reciprocal percentile-rank agreement across 21 model pairs, while the right panel presents the 15 most consistently close task pairs using mean within-model-pair percentile and top-decile prevalence. Since raw MMD values are not comparable across SAE model pairs, all results are rank-normalized within each model pair before aggregation.
Same-family task pairs also rank 3.1 percentile points worse than cross-family pairs on average, with the same trend in 16 of 21 model pairs.

\vspace{-3pt}
\begin{tcolorbox}[
    takeaways,
    title=Key Takeaway IV,
    colbacktitle=cyan!60!black
]
Despite the difficulty of cross-task, cross-model comparison, a small set of specific task/model combinations still exhibits consistently low activation-rate MMD under the joint screening criteria. These similarities are pair-specific, and same-family task pairs are not closer on average.
\end{tcolorbox}
\vspace{-5pt}

\subsection{Discussion}
\label{sec:discussion}

Our results reveal meaningful geometric organization of low-level visual tasks (RQ1). Within each model, task instructions induce consistent global displacement directions across images, while LID follows a similar task-dependent ordering across architectures.
This organization is also interpretable and partly shared at the feature level (RQ2): sparse features are broadly reused across tasks within each model, and the inter-task geometry is better explained by shared degradation and transformation mechanisms than by the predefined perceptual families.
Overall, global similarity and local geometry provide complementary views of a mechanism-centric task structure.
Cross-model universality is substantial but limited (RQ3). Different VLMs preserve similar global distance relationships for the same task, but exhibit weaker agreement in local neighborhoods and coordinate alignment. This suggests that universality emerges mainly through shared relational organization rather than pointwise feature correspondence. Sparse feature analysis may reveal recurring visual primitives, but feature-level sharing should be established through explicit cross-model validation.

\vspace{-8pt}
\paragraph{Limitations.}
Our study has several limitations:
(i) analyzing a single selected layer per model may overlook layer-specific geometric and sparse-feature dynamics;
(ii) standardized prompts together with a transformer-dominated model set may limit generalization to more diverse instructions and architectures;
(iii) task comparisons may also reflect dataset differences, since different tasks draw on different source datasets;
and (iv) projection-based topology and cross-model SAE findings require further validation across layers, methods, and model pairs.

\section{Conclusion}
\label{sec:conclusion}

In this study, we present a comprehensive geometric analysis of how VLMs represent low-level visual tasks and the relations among cross-task representations. We propose GeoSim, a four-level analysis framework comprising global similarity, local geometry, sparse feature decomposition, and topological verification.
This work establishes a benchmark for future VLM research on representation interpretability, transfer learning, and representation alignment across low-level tasks. More importantly, our analysis provides actionable insights into how low-level task representations are organized in VLMs, offering guidance for the development of VLMs in all-in-one image restoration.

\bibliography{ref}
\bibliographystyle{iclr2027_conference}

\clearpage
\typeout{>>> Appendix starts at physical page \the\numexpr\ReadonlyShipoutCounter+1\relax}
\setcounter{page}{1}
\ifdefined\iclrrulercount \global\iclrrulercount=0 \fi
\appendix
\section{Extended Related Work}
\label{app:related}

\subsection{Image Restoration and Task Conditioning}

Early approaches primarily relied on shared CNN- or Transformer-based representations to model multiple degradation types, including AirNet~\citep{li2022all} and TransWeather~\citep{valanarasu2022transweather}. IDR~\citep{zhang2023ingredient} introduced an ingredient-oriented formulation that explicitly disentangles degradation factors into reusable restoration components.
Subsequent methods further improved task conditioning by incorporating prompts or natural-language instructions. PromptIR~\citep{potlapalli2023promptir} employs learnable prompts to adapt a unified restoration backbone to different degradations, while InstructIR~\citep{conde2024instructir} replaces task-specific embeddings with human-readable language instructions.

\subsection{Task Transferability}

Taskonomy was subsequently extended to zero-shot task transfer~\citep{pal2019tmt}.
Transferability has also been studied for VLMs: \citet{chen2024learning} investigated knowledge transferability across multimodal tasks, while \citet{tiong2024we} analyzed latent factors underlying VLM benchmarks and showed that benchmark performance may reflect evaluation biases rather than genuine task competence.

\subsection{Geometric Representation Analysis}

Geometric representation analysis seeks to understand neural networks by characterizing the structure of their hidden representations.
Beyond the analyses discussed in Section~\ref{sec:related}, existing geometric analysis methods can be broadly divided into four categories: (1) global similarity methods, including RSA~\citep{kriegeskorte2008representational}, CKA~\citep{kornblith2019similarity}, and dCor~\citep{szekely2007measuring}, quantify representational similarity between activation spaces; (2) alignment methods, such as Procrustes analysis~\citep{williams2021generalized}, explicitly compare geometric structures across models; (3) manifold-based analyses estimate intrinsic geometric properties through measures such as LID~\citep{levina2004maximum}; and (4) distributional approaches, including MMD~\citep{gretton2012kernel}, compare activation distributions.
Topological RSA~\citep{lin2024topology} further provides a controlled continuum between complete distance geometry and neighborhood structure.

\subsection{Sparse Autoencoders Across Domains}

Early work showed that SAEs can recover approximately monosemantic features from language-model activations~\citep{huben2024sparse}. Subsequent studies established scalable training and evaluation protocols~\citep{gao2025scaling} and demonstrated that learned sparse features can support interpretation and causal steering~\citep{templeton2026scaling}.

More recently, SAE-based analysis has been extended to vision and vision-language models.
SpLiCE~\citep{bhalla2024interpreting} decomposes CLIP embeddings into sparse language-grounded concepts, while PatchSAE~\citep{lim2025sparse} applies SAEs to patch-level visual activations to identify localized concepts and studies how prompt adaptation remaps existing visual concepts.

Later work showed that interpretable SAE features are not necessarily steerable~\citep{kulkarni2026interpretable}, and that semantic labels alone do not reveal whether a visual feature uses local or global evidence~\citep{ro2026beyond}.
Other recent studies also extend sparse representations across modalities by transferring language-side interpretations to vision encoders~\citep{kravets2026interpretability}, extracting unified visual, linguistic, and multimodal concepts within a single VLM~\citep{lanza2026extraction}, and jointly learning visual and textual sparse features for behavioral steering~\citep{shu2026steering}.

\section{Additional Framework Details}
\label{app:framework}

\subsection{Overview of the Analysis Levels}

GeoSim analyzes representations from four complementary perspectives that progressively reveal different aspects of their organization:

\begin{enumerate}
    \item \textbf{Global Similarity Structure:}
    How are different low-level visual tasks organized in representation
    space, and how does this organization evolve across network depth?

    \item \textbf{Local Geometric Structure:}
    Do tasks with similar global representations also preserve comparable
    neighborhood structures and manifold properties?

    \item \textbf{Sparse Feature Decomposition:}
    Which latent features are shared across tasks, and which features
    encode task-specific information?

    \item \textbf{Cross-model Universality:}
    To what extent do different VLM architectures develop similar
    geometric structures for the same perceptual tasks?
\end{enumerate}

Together, these four levels provide a hierarchical characterization of low-level visual representations, progressing from task-level organization to local geometry, feature-level mechanisms, and cross-model consistency. The notation used throughout the framework is summarized in Table~\ref{tab:notation}.

\subsection{Motivation and Complementarity of the Analysis Levels}
\label{app:level-roles}

\subsubsection{Global and Local Geometry}

Level~1 investigates the global organization of low-level task representations. We first quantify the similarity between different tasks within each VLM to determine whether perceptually related tasks form coherent representation clusters. We then evaluate whether the resulting global structures are consistent across different VLM architectures and how they evolve throughout the network.

Level~1 characterizes the global relationships among low-level tasks but cannot determine whether these similarities are preserved locally. Level~2 therefore examines neighborhood structure and manifold geometry to distinguish genuine geometric correspondence from global averaging effects.

While Level~1 captures global task organization through aggregate similarity, it cannot show whether these relationships hold at local scales. Level~2 addresses this gap by examining three complementary properties of representation spaces, namely local manifold complexity, neighborhood consistency, and coordinate-invariant geometric alignment. Together, these measures reveal whether global task similarity reflects genuine structural correspondence or merely high-dimensional averaging.

Level~2 complements the global similarity analysis in Level~1 by testing whether task relationships persist at finer geometric scales. High global similarity accompanied by consistent neighborhood structure and alignment provides stronger evidence of shared representation geometry, whereas discrepancies indicate that global correlations may obscure local differences.

\subsubsection{Sparse Features and Topological Verification}

Levels~1--2 characterize representation geometry through continuous metrics, including similarity, neighborhood structure, and alignment. Level~3 moves beyond geometric correspondence to identify the sparse latent features associated with shared and task-specific representation structures. Specifically, we decompose hidden representations using SAEs and analyze task-dependent feature activation patterns as an interpretable complement.

Level~3 provides a feature-level interpretation of the geometric patterns identified in Levels~1--2, but depends on SAE training choices such as dictionary size and sparsity regularization. Level~4 therefore examines topological properties of representation spaces using geometric invariants, providing an additional analysis that is less dependent on a particular feature decomposition.

While Levels~1--3 analyze representation space through continuous geometry and sparse SAE activations, they remain sensitive to metric and dictionary choices. Level~4 introduces a complementary topological perspective using persistent homology (PH). Because persistence diagrams are invariant to isometric transformations of the point cloud, Level~4 does not depend on a shared coordinate frame; it still depends on projection and filtration choices, and thus provides a cross-check under different assumptions rather than an assumption-free refinement.

Level~4 does not replace the geometric and feature-level analyses in Levels~1--3. Instead, it evaluates whether their conclusions remain consistent when representation structure is summarized through topological invariants. Agreement across levels provides stronger evidence that observed task organization reflects intrinsic properties of VLM representations rather than artifacts of a particular metric, feature decomposition, or analysis method.

\subsection{Additional Analysis Metrics}

\subsubsection{Level 1: Global Similarity Structure}
\label{app:dcor-details}

\paragraph{Cross-model similarity.}
For the dCor statistic in Eq.~\eqref{eq:dcor}, pairwise Euclidean distances are computed as
\begin{equation}
a_{ij}^{s}
=
\|h_t^{\ell,M_1}(x_i^s)-h_t^{\ell,M_1}(x_j^s)\|_2,
\qquad
b_{ij}^{s}
=
\|h_t^{\ell,M_2}(x_i^s)-h_t^{\ell,M_2}(x_j^s)\|_2,
\label{eq:dcor-distance}
\end{equation}
and double-centered to obtain
$A_{ij}^{s}
=
a_{ij}^{s}
-
\bar a_{i\cdot}^{s}
-
\bar a_{\cdot j}^{s}
+
\bar a_{\cdot\cdot}^{s}$
and
$B_{ij}^{s}
=
b_{ij}^{s}
-
\bar b_{i\cdot}^{s}
-
\bar b_{\cdot j}^{s}
+
\bar b_{\cdot\cdot}^{s}$.
The squared distance covariance and variances are then computed as
\begin{equation}
\big(\mathrm{dCov}^{s}\big)^{2}=\frac{1}{n^{2}}\sum_{i,j}A_{ij}^{s}B_{ij}^{s},\quad
\big(\mathrm{dVar}_1^{s}\big)^{2}=\frac{1}{n^{2}}\sum_{i,j}\big(A_{ij}^{s}\big)^{2},\quad
\big(\mathrm{dVar}_2^{s}\big)^{2}=\frac{1}{n^{2}}\sum_{i,j}\big(B_{ij}^{s}\big)^{2}.
\label{eq:dcor-statistics}
\end{equation}

\subsubsection{Level 2: Local Geometric Structure}
\label{app:lid}

\paragraph{Local intrinsic dimensionality (\lid).}
We first characterize the local complexity of each task representation space using \lid, estimated by maximum likelihood~\citep{levina2004maximum}.
For task $t$ at layer $\ell$, the \lid{} of sample $x_i^s$ in sampling repeat $s$ is
\begin{equation}
\widehat{\lid}_k^\ell(x_i^s;t)
=
\left[
\frac{1}{k-1}
\sum_{o=1}^{k-1}
\log
\frac{
r_k(h_t^\ell(x_i^s))
}{
r_o(h_t^\ell(x_i^s))
}
\right]^{-1},
\label{eq:lid}
\end{equation}
where $r_o(h_t^\ell(x_i^s))$ denotes the Euclidean distance from $h_t^\ell(x_i^s)$ to its $o$-th nearest neighbor among samples from the same task representation space. The average \lid{} of task $t$ is obtained by averaging over samples and sampling repeats:
\begin{equation}
\overline{\lid}^{\,\ell}(t)
=
\frac{1}{b}
\sum_{s=1}^{b}
\frac{1}{n}
\sum_{i=1}^{n}
\widehat{\lid}_k^\ell(x_i^s;t).
\label{eq:mean_lid}
\end{equation}

\subsubsection{Level 3: Sparse Feature Decomposition}
\label{app:level3-details}

\paragraph{SAE architecture.}
At the stable layer $\ell^*$, we denote the extracted representation as $z_t(x)=h_t^{\ell^*}(x)\in\mathbb{R}^{d}$.
For each VLM, we train a single SAE shared across all low-level vision tasks, allowing feature usage to be compared within a common learned dictionary. The SAE encoder maps the representation into a sparse latent code
\begin{equation}
a(x)=\mathrm{ReLU}(W_e z_t(x)+b_e),
\label{eq:sae-encoder}
\end{equation}
where $W_e\in\mathbb{R}^{D_{\mathrm{SAE}}\times d}$ and $b_e\in\mathbb{R}^{D_{\mathrm{SAE}}}$ are the encoder parameters, $a(x)\in\mathbb{R}^{D_{\mathrm{SAE}}}$ is the sparse feature activation vector, and $D_{\mathrm{SAE}}$ denotes the SAE dictionary size.

The original representation is reconstructed by
\begin{equation}
\hat z_t(x)=W_d a(x)+b_d,
\label{eq:sae-decoder}
\end{equation}
where $W_d\in\mathbb{R}^{d\times D_{\mathrm{SAE}}}$ and $b_d\in\mathbb{R}^{d}$ are the decoder parameters. The SAE is optimized using a reconstruction loss with an $\ell_1$ sparsity penalty:
\begin{equation}
\mathcal{L}_{\mathrm{SAE}}
=
\mathbb{E}_{x}
\left[
\frac{1}{d}
\|z_t(x)-\hat z_t(x)\|_2^2
+
\frac{\lambda}{D_{\mathrm{SAE}}}
\|a(x)\|_1
\right].
\label{eq:sae-loss}
\end{equation}

\vspace{-8pt}
\paragraph{Task-pair feature sharing.}
While $I(Z_f;T)$ in Eq.~\eqref{eq:mi-feature-task} measures whether individual features encode task identity, it does not quantify whether two tasks rely on shared sparse features. Therefore, for task pairs with shared underlying image identities, we measure the similarity between their binary SAE activation patterns. Let $F_p$ and $F_q$ denote the task-wise binary SAE activation
patterns for tasks $p$ and $q$, respectively. Their dependency is
quantified using \mi:
\begin{equation}
I(F_p;F_q)
=
\sum_{c\in\{0,1\}}
\sum_{e\in\{0,1\}}
P_{pq}(c,e)
\log
\frac{P_{pq}(c,e)}
{P_p(c)P_q(e)},
\label{eq:task-pair-mi}
\end{equation}
where $c$ and $e$ denote the binary activation states associated with
tasks $p$ and $q$, respectively. Here, $P_{pq}(c,e)$ is their joint
probability, while $P_p(c)$ and $P_q(e)$ are the corresponding marginal
probabilities, estimated from paired samples. Higher values indicate stronger feature-level overlap between two tasks, providing an information-theoretic complement to the geometric similarity measured in Levels~1--2.

\subsubsection{Level 4: Topological Verification}
\label{app:topology-details}

\paragraph{Persistent homology.}
The point cloud $\widetilde{\HH}_t$ is obtained using the UMAP preprocessing described in Section~\ref{sec:level4}.

Given $\widetilde{\HH}_t$, we construct the Vietoris--Rips filtration~\citep{zomorodian2005computing}:
\begin{equation}
\mathrm{VR}_{\varepsilon}(\widetilde{\HH}_t)
=
\left\{
\sigma\subseteq\widetilde{\HH}_t:
\operatorname{diam}(\sigma)\leq\varepsilon
\right\},
\qquad \varepsilon\geq0 ,
\label{eq:vr}
\end{equation}
which forms a nested sequence of simplicial complexes as $\varepsilon$ increases. At each filtration scale, the $m$-th homology group
$\mathcal{G}_m(\mathrm{VR}_{\varepsilon}(\widetilde{\HH}_t))$
captures topological structures of dimension $m$, summarized by the Betti number
\begin{equation}
\beta_m(\varepsilon)
=
\operatorname{rank}
\mathcal{G}_m
\left(
\mathrm{VR}_{\varepsilon}(\widetilde{\HH}_t)
\right).
\label{eq:betti}
\end{equation}
We focus on $\beta_0$, corresponding to connected components, and $\beta_1$, corresponding to one-dimensional loops.
Following common usage, we refer to degrees $m = 0, 1$ as $H_0$ and $H_1$.
Each topological feature appears at a birth scale $\alpha$ and disappears at a death scale $\delta>\alpha$. The resulting birth--death pairs form the persistence diagram:
\begin{equation}
\mathrm{PD}_m(\widetilde{\HH}_t)
=
\left\{
(\alpha_1,\delta_1),
(\alpha_2,\delta_2),
\ldots
\right\}.
\label{eq:pd}
\end{equation}

\vspace{-8pt}
\paragraph{Cross-model topological consistency.}
To quantify whether different VLMs produce similar topological structures, we compare persistence diagrams from models $M_1$ and $M_2$ for the same task $t$. Specifically, we compute the $r$-Wasserstein distance:
\begin{equation}
d_W
\left(
\mathrm{PD}_m^{M_1},
\mathrm{PD}_m^{M_2}
\right)
=
\left(
\inf_{\gamma}
\sum_{\xi\in\mathrm{PD}_m^{M_1}}
\left\|
\xi-\gamma(\xi)
\right\|_\infty^r
\right)^{1/r},
\label{eq:wasserstein}
\end{equation}
where $\gamma$ denotes a matching between persistence diagrams, with diagonal points added to handle unequal numbers of features.
As $r\rightarrow\infty$, the Wasserstein distance reduces to the bottleneck distance:
\begin{equation}
d_B
\left(
\mathrm{PD}_m^{M_1},
\mathrm{PD}_m^{M_2}
\right)
=
\inf_{\gamma}
\sup_{\xi\in\mathrm{PD}_m^{M_1}}
\left\|
\xi-\gamma(\xi)
\right\|_\infty.
\label{eq:bottleneck}
\end{equation}
We report the Wasserstein distance with $r=2$ as the primary metric because it captures the aggregate deformation of persistence diagrams, while the bottleneck distance is reported as a complementary worst-case measure.
We additionally assess sampling variability for both Conditions II and IV: in Figure~\ref{fig:figure2_2_Appendix}, intervals show the spread across the $b=10$ frozen sampling repeats, whereas in Figure~\ref{fig:figure4_Appendix}, they are bootstrap percentile intervals from $B_{\mathrm{PH}}=50$ resamples.

\vspace{-8pt}
\paragraph{Stability considerations.}
The stability of persistence diagrams under perturbations of the input point cloud has been established for the bottleneck distance~\citep{cohensteiner2007stability} and, under additional Lipschitz and bounded-total-persistence assumptions, extended to finite $r$-Wasserstein distances~\citep{cohensteiner2010lipschitz}. These results justify using persistence-based distances as robust summaries of geometric structure, but do not eliminate sensitivity introduced by dimensionality reduction or filtration choices.

\section{Data Details}
\label{app:data_details}

\subsection{Datasets}
This section provides additional details about the datasets used to construct the 24 low-level vision tasks in our experiments.

\subsubsection{Restoration}

\paragraph{LIU4K.}
LIU4K~\citep{liu2020liu4k} is a high-resolution dataset for single-image compression artifact removal, containing 1,500 training images and 200 testing images at $4$K resolution. Since it provides only clean images, we generate compressed inputs using the compression settings adopted in our experiments and retain the original images as references.

\vspace{-8pt}
\paragraph{DPDD.}
DPDD~\citep{abuolaim2020defocus} is a real-world defocus deblurring dataset captured using a Canon EOS 5D Mark IV dual-pixel camera. It contains 500 scenes and 2,000 images. Each scene provides a defocused image, left and right dual-pixel sub-aperture views, and an all-in-focus reference image. The official split consists of 350 training scenes, 74 validation scenes, and 76 testing scenes.

\vspace{-8pt}
\paragraph{GoPro.}
GoPro~\citep{nah2017deep} is a motion deblurring dataset in which blurry images are synthesized by averaging consecutive high-frame-rate frames, with the corresponding sharp frames used as references. It contains 3,214 blurry and sharp pairs at a resolution of $1280\times720$, including 2,103 pairs from 22 sequences for training and 1,111 pairs from 11 sequences for testing.

\vspace{-8pt}
\paragraph{D-HAZY.}
D-HAZY~\citep{ancuti2016dhazy} is a synthetic dehazing benchmark that generates hazy images from haze-free RGB images and their associated depth maps using the atmospheric scattering model, producing aligned hazy and haze-free pairs. We use its NYU Depth v2 subset, generated from the 1,449 images of NYU Depth v2~\citep{silberman2012indoor}, which cover 464 indoor scenes captured using a Microsoft Kinect camera.

\vspace{-8pt}
\paragraph{UHDM.}
UHDM~\citep{yu2022towards} is a real-world $4$K image demoir\'eing dataset containing 5,000 paired moir\'e and moir\'e-free images, including 4,500 pairs for training and 500 pairs for testing. The images were collected using different display devices and cameras and cover moir\'e patterns with diverse colors, orientations, frequencies, and spatial scales.

\vspace{-8pt}
\paragraph{SIDD.}
SIDD~\citep{abdelhamed2018high} is a real-world smartphone denoising dataset containing approximately 30,000 noisy images from 10 representative scenes captured using five smartphone cameras. The images cover different illumination conditions, ISO levels, exposure times, and scene contents. Corresponding nearly noise-free reference images are obtained from repeated captures of the same scenes.

\vspace{-8pt}
\paragraph{Rain14000.}
Rain14000~\citep{fu2017removing} is a synthetic deraining dataset constructed from 1,000 clean images. Fourteen rainy images are generated for each clean image using rain patterns with different orientations and intensities, resulting in 14,000 rainy and clean pairs. The standard split contains 12,600 pairs generated from 900 clean images for training and 1,400 pairs generated from the remaining 100 clean images for testing.

\vspace{-8pt}
\paragraph{Snow100K.}
Snow100K~\citep{liu2018desnownet} is a synthetic desnowing dataset containing 100,000 snowy images produced by compositing snow masks onto snow-free photographs, split evenly into 50,000 training and 50,000 testing images. Each synthesized image is accompanied by its snow-free reference and the corresponding snow mask, and the test set is further partitioned into light, medium, and heavy subsets according to snow particle size, so that the benchmark covers snow particles with different sizes, densities, and transparencies. The dataset additionally provides a set of realistic snowy photographs without references. We use the paired synthetic images, taking the snowy image as the input and the snow-free image as the reference.

\vspace{-8pt}
\paragraph{UIEB.}
UIEB~\citep{li2019underwater} is a real-world underwater image enhancement benchmark containing 950 images with diverse color casts, visibility conditions, illumination patterns, and underwater environments. Among them, 890 images have reference images selected through human evaluation, while the remaining 60 challenging images do not have reference targets. We use the paired subset for underwater restoration.

\subsubsection{Removal}

\paragraph{Flare7K++.}
Flare7K++~\citep{dai2024flare7kpp} is a large-scale lens flare removal dataset that extends Flare7K by combining 7,000 synthetic flare images, including 5,000 scattering flares and 2,000 reflective flares, with 962 real captured flare images and 23,949 flare-free background images. Flare-corrupted inputs in the released paired test set are generated by compositing flare layers with clean backgrounds, while the original backgrounds serve as references.

\vspace{-8pt}
\paragraph{Raindrop Clarity.}
Raindrop Clarity~\citep{jin2024raindrop} is a real-world raindrop removal dataset covering both daytime and nighttime conditions. It contains 15,186 image pairs or triplets, including 5,442 daytime samples and 9,744 nighttime samples. The dataset provides raindrop-corrupted images, clear background images, and observations captured under different focal settings. It includes both background-focused cases, in which raindrops appear blurred, and raindrop-focused cases, in which the background may be out of focus.

\vspace{-8pt}
\paragraph{SIR$^2$.}
SIR$^{2}$~\citep{wan2017benchmarking} is a real-world benchmark for single-image reflection removal. It contains 40 controlled indoor scenes, including 20 solid-object scenes and 20 postcard scenes, together with 100 wild scenes. The controlled scenes provide reflection-contaminated images and their corresponding transmission and reflection layers. Multiple observations are captured using different aperture settings and glass thicknesses, producing reflections with different strengths, blur levels, and spatial structures.

\vspace{-8pt}
\paragraph{ISTD.}
ISTD~\citep{wang2018stacked} is a shadow removal dataset containing 1,870 image triplets, each consisting of a shadow image, its corresponding shadow-free image, and a binary shadow mask. The official split contains 1,330 triplets for training and 540 triplets for testing. The dataset covers shadows with different shapes, strengths, and boundary characteristics.

\subsubsection{Generation and Enhancement}

\paragraph{ADE20K.}
ADE20K~\citep{zhou2017scene} is a scene-understanding dataset covering diverse indoor and outdoor scenes, objects, materials, and spatial layouts. Its standard scene-parsing benchmark contains 20,210 training images, 2,000 validation images, and 3,000 testing images, with annotations for 150 semantic categories. For colorization, we use the RGB images rather than their semantic annotations. Each RGB image is converted to grayscale as input, while the original color image is retained as the reference.

\vspace{-8pt}
\paragraph{iHarmony4.}
iHarmony4~\citep{cong2020dovenet} is a large-scale image harmonization benchmark combining four sub-datasets: HCOCO, HAdobe5K, HFlickr, and Hday2night. It contains 73,146 synthesized composite images, each accompanied by a harmonized reference image and a foreground mask. The composite foregrounds exhibit inconsistencies with their backgrounds in color, illumination, contrast, and overall appearance.

\vspace{-8pt}
\paragraph{Natural Images.}
The Natural Images dataset, proposed by \citet{roy2018effects}, contains 6,899 high-resolution natural images from eight object categories, including airplane, car, cat, dog, flower, fruit, motorbike, and person. It is split into 5,724 training images and 1,175 validation images. Since the dataset provides only intact images, we generate masked images and use the original images as ground-truth targets for image inpainting.

\vspace{-8pt}
\paragraph{LOL.}
LOL~\citep{wei2018deep} is a real-world low-light image enhancement dataset containing 500 aligned low-light and normal-light image pairs captured from the same scenes under different exposure conditions. The standard split contains 485 pairs for training and 15 pairs for testing. The low-light images exhibit reduced visibility, color distortion, amplified noise, and loss of local details.

\vspace{-8pt}
\paragraph{Night2Day.}
The Night2Day dataset distributed with pix2pix~\citep{isola2017image} is derived from the Transient Attributes Database~\citep{laffont2014transient}. It contains approximately 20,000 natural-scene images captured by stationary webcams at different times of day. Nighttime and daytime observations of the same underlying scenes are used as source and reference images for style transfer.

\vspace{-8pt}
\paragraph{BSDS500.}
BSDS500~\citep{arbelaez2011contour} is a standard benchmark for natural-image edge and boundary detection. It contains 500 images, divided into 200 training images, 100 validation images, and 200 testing images. Each image is annotated by multiple human subjects, resulting in several valid boundary maps representing perceptually meaningful object and region contours.

\subsubsection{Reconstruction}

\paragraph{DIV2K.}
DIV2K~\citep{agustsson2017ntire} is a high-quality image super-resolution dataset containing 1,000 images at approximately $2$K resolution, including 800 training images, 100 validation images, and 100 testing images. The corresponding low- and high-resolution images are used as inputs and references for the super-resolution task.

\vspace{-8pt}
\paragraph{HDRTV1K.}
HDRTV1K~\citep{chen2021new} is an ultra-high-definition dataset for SDRTV-to-HDRTV conversion. It contains 1,235 training pairs and 117 testing pairs at $4$K resolution, derived from HDR10 video content. The paired SDRTV and HDRTV images differ in dynamic range, color gamut, contrast, and highlight appearance, with SDRTV images serving as source observations and HDRTV images as references.

\subsubsection{Photometric Correction}

\paragraph{Rendered WB.}
The Rendered WB dataset~\citep{afifi2019when} is a large-scale benchmark for white-balance correction. Its main set contains 62,535 rendered images captured using seven DSLR cameras, with corresponding correctly white-balanced references. An additional cross-camera test set contains 2,881 rendered images captured using four mobile phones and one DSLR camera. Each scene is rendered under several camera-specific white-balance settings to produce realistic color casts.

\vspace{-8pt}
\paragraph{SICE.}
SICE~\citep{cai2018learning} is a multi-exposure contrast enhancement dataset containing 589 indoor and outdoor scene sequences and 4,413 images in total. Each sequence depicts the same static scene under several exposure levels. Candidate enhanced images are generated using multiple multi-exposure fusion and HDR methods, after which subjective evaluation is used to select a high-quality reference image for each sequence.

\vspace{-8pt}
\paragraph{VIDIT.}
For relighting, we use Track 1 of the AIM 2020 Scene Relighting Challenge~\citep{elhelou2020aim}, based on VIDIT~\citep{elhelou2020vidit}. VIDIT contains 390 synthetic scenes at a resolution of $1024\times1024$, each rendered under 40 illumination settings formed by eight illumination directions and five color temperatures. The scenes are divided into 300 training scenes, 45 validation scenes, and 45 testing scenes. Track 1 follows a one-to-one relighting setting, in which an image under a source illumination is transformed into the same scene under a predefined target color temperature and illumination direction.

\begin{figure*}[t]
    \centering
    \includegraphics[width=\textwidth]{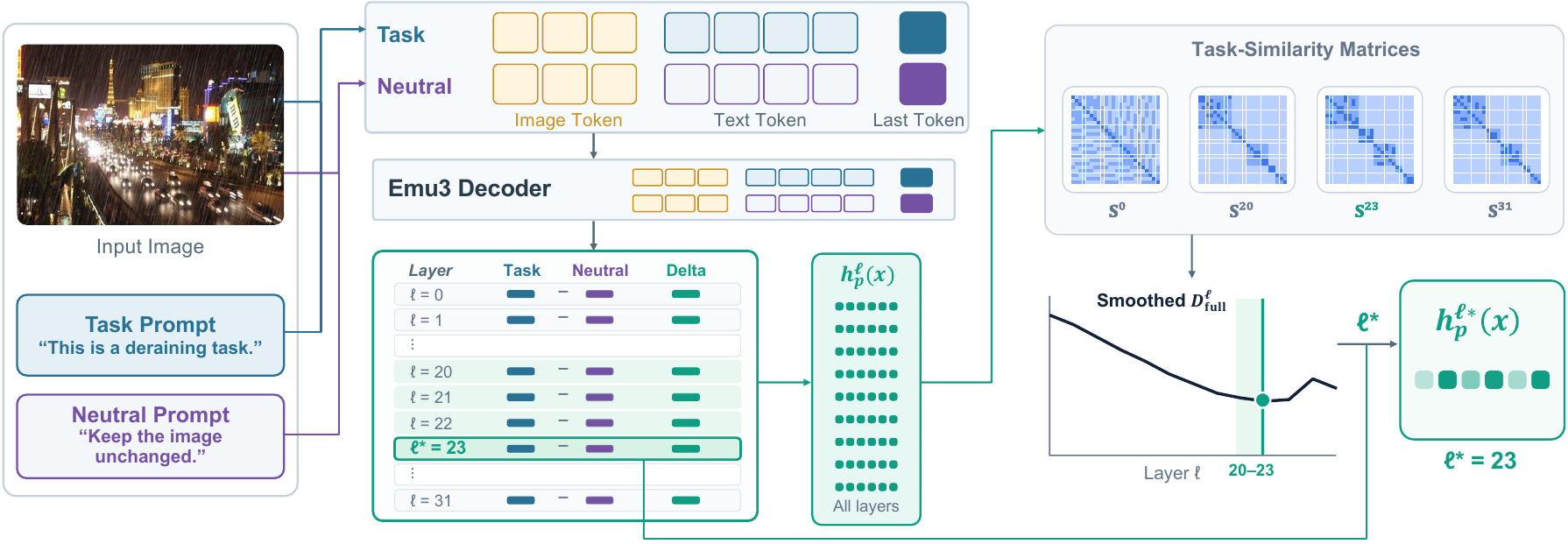}
    \caption{\textbf{Feature extraction and stable-layer selection.}
    Illustrated using a typical AR model (Emu3). Input images with task and neutral prompts produce the cached states whose difference defines $h_p^{\ell}(x)$.}
    \label{fig:feature-extraction-flow}
    \vspace{-10pt}
\end{figure*}

\subsection{Feature Extraction Protocol}
\label{app:extraction}

All images are EXIF-oriented, converted to RGB, and resized to $512\times512$. For each image $x$ and task $p$, the frozen model is run with the fixed task instruction ``This is a/an [task name] task'' and with the neutral instruction ``Keep the image unchanged'' as the baseline. For AR models, the attention mask identifies the final valid prompt position, which can attend to both the image and the complete instruction. We use the difference between its task and neutral states,
\begin{equation}
h_p^{\ell}(x)
=u_{\mathrm{task},\mathrm{last}}^{\ell}(x)
-u_{\mathrm{neutral},\mathrm{last}}^{\ell}(x).
\label{eq:instruction-delta-last}
\end{equation}
For Qwen-Image-Edit, a DiT model, the corresponding representation is the task-minus-neutral difference between spatial means of the condition-image feature maps at the first denoising step.

Figure~\ref{fig:feature-extraction-flow} illustrates how we extract hidden-layer representations, using a typical AR model as an example.
At each layer of the AR model, we cache the last-token hidden states for the task-prompt and neutral-prompt input sequences, along with the sample IDs and extraction metadata. Eq.~\eqref{eq:instruction-delta-last} is applied when the cache is loaded. Each frozen manifest, a saved ordered list of sample IDs reused across models, selects $n$ rows in its recorded order.

To choose the stable layer $\ell^*$, we average the $b$ task-similarity matrices per layer, compute
\begin{equation}
D_{\mathrm{full}}^{\ell}
=
\left\|
\mathbf{S}^{\ell}
-
\mathbf{S}^{\ell+1}
\right\|_F.
\label{eq:frobenius-evolution}
\end{equation}
We smooth the change curve with a three-transition rolling median. Here, ``full'' denotes the complete task-similarity matrix. The task margin is the mean within-task similarity minus the mean cross-task similarity. Within 15--95\% of network depth, the stable range begins at the first two-transition run below the 0.35-quantile change threshold whose task margin is at least the median candidate margin. The range ends when the change threshold is exceeded, and its final layer is used for all single-layer geometric analyses. If no candidate meets both criteria, the layer with the highest combined normalized margin and inverse-change score is selected and flagged as a fallback.
SAE layer selection uses the separate resampling protocol in Appendix~\ref{app:sae}.

\subsection{SAE Training Details}
\label{app:sae}

For the SAE analyses in Conditions~II and IV, stable-layer selection uses 50 within-task bootstrap resamples of 100 rows, drawn with replacement from a 50\% selection subset, and applies the criterion in Appendix~\ref{app:extraction}. These resamples are distinct from the $b=10$ frozen sampling repeats and the $B_{\mathrm{PH}}=50$ topological resamples.

We train one SAE per model on the $\ell^*$ representations from all 24 tasks. Samples are first split by base-content identity into 70/30 training and evaluation sets, before any balancing or normalization. In the paired-control setting, identical samples are kept in the same split across tasks. Each task contributes $n_{\mathrm{SAE}}=500$ training samples, with replacement used only when necessary. The training-set mean and scalar root-mean-square deviation are then used to apply the same normalization to both splits.

The trained SAE with a ReLU activation function and an $\ell_1$ penalty in Eqs.~\eqref{eq:sae-encoder}--\eqref{eq:sae-loss} uses $D_{\mathrm{SAE}}=8d$. For each model, $\lambda$ is selected on an inner 80/20 split from $\{0.1,0.3,1,3,10,30\}$ after 2,000 steps per candidate. Eligible candidates require explained variance $\geq 0.80$ and active fraction $\leq 0.05$. Among these, we choose the one closest to the target active fraction 0.02, breaking ties by explained variance and then smaller $\lambda$. Final fits use seeds $\{0,1,2\}$ and AdamW for 5,000 steps (learning rate $10^{-3}$, batch size 1,024, zero weight decay, gradient clipping 1.0, and unit-norm decoder columns). All seeds must also satisfy the stated dead-feature, cross-seed, and train--evaluation checks, including explained variance $\geq 0.80$, active fraction $\leq 0.05$, dead-feature fraction $\leq 0.95$, cross-seed explained-variance and active-fraction ranges $\leq 0.05$ and $\leq 0.02$, and train--evaluation explained-variance gap $\leq 0.15$.

Level-3 metrics use only held-out codes and define a feature as active when $a_f>\tau=10^{-6}$. Among features active on at least one held-out sample, we label a feature task-specific if $\tsi(f)>\theta_{\mathrm{high}}=0.85$, shared if $\tsi(f)<\theta_{\mathrm{low}}=0.30$, and intermediate otherwise. Crucially, metrics such as TSI, MI, and MMD are not used during $\lambda$ selection.

\begin{figure*}[t]
    \centering
    \includegraphics[width=\textwidth]{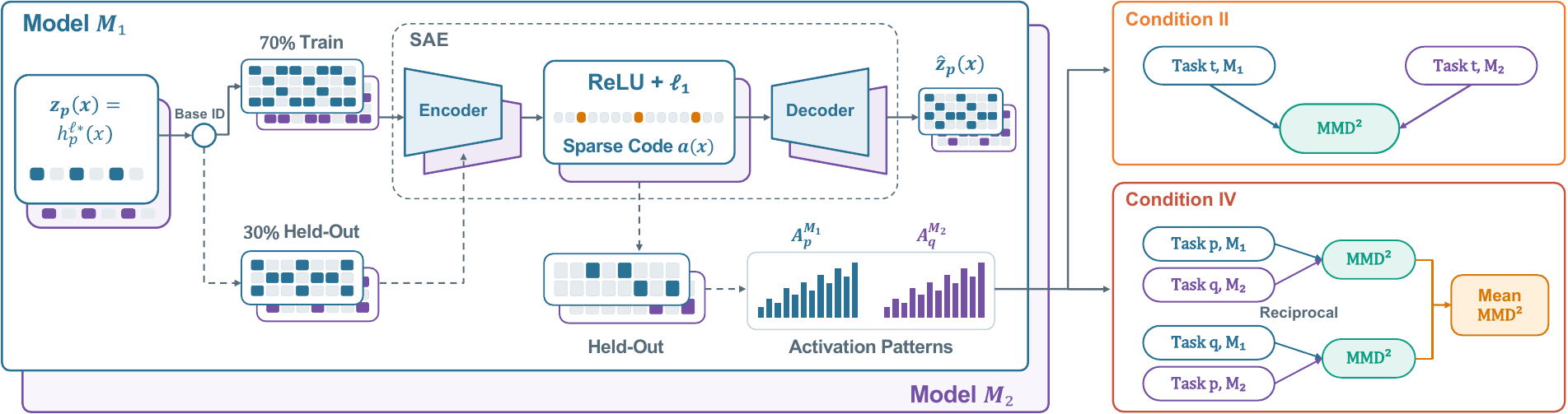}
    \caption{\textbf{SAE training and its use in Conditions II and IV\@.}
    Per-model SAEs are fit on task-balanced training data and reused to compare
    held-out activation-rate distributions. Neither condition assumes aligned
    SAE feature coordinates.}
    \label{fig:sae-training-conditions}
    \vspace{-10pt}
\end{figure*}

Conditions II and IV reuse these per-model SAEs and evaluation codes. Because their feature coordinates are unaligned, cross-model MMD compares the unordered activation-rate distributions rather than raw coordinates. The RBF kernel uses a bandwidth equal to the square root of the median positive squared pairwise distance among the pooled activation rates in each comparison.
Condition II uses different models with the same task. Condition IV uses different models and tasks, averages the two reciprocal task assignments, and ranks the result within each model pair. Figure~\ref{fig:sae-training-conditions} illustrates the SAE training process and its use in both conditions.

\section{Supporting Materials}

\subsection{Mapping Between Conditions and Analysis Levels}

Table~\ref{tab:map} maps the four evaluation conditions in Section~\ref{sec:result} to the corresponding GeoSim analysis level(s) and the associated evaluation metrics.
Overall, the table serves as a roadmap for the experimental analysis by linking each evaluation condition to its analysis objective and the appropriate metrics, providing an overview of the complete evaluation protocol.

\begin{table*}[t]
\centering
\caption{Task/model conditions, and corresponding analysis levels and metrics.}
\label{tab:map}
\scriptsize
\setlength{\tabcolsep}{2.5pt}
\renewcommand{\arraystretch}{1.3}
\begin{tabular}{@{}
>{\raggedright\arraybackslash}p{1.75cm}
>{\raggedright\arraybackslash}p{2.9cm}
>{\raggedright\arraybackslash}p{2.05cm}
>{\raggedright\arraybackslash}p{6.6cm}@{}}
\toprule
Condition & Setup & Level(s) & Metrics \\
\midrule

\makecell[tl]{Same Task, \\Same Model}
& One task and model, with images varied
& \makecell[tl]{L1 (primary)\\L2 (supplementary)}
& Intra-task cosine (Eq.~\eqref{eq:intra-task-pair-cosine}),
  Frobenius distance (Eq.~\eqref{eq:frobenius-evolution}),
  \newline LID (Eqs.~\eqref{eq:lid}--\eqref{eq:mean_lid}) \\

Same Task, Different Models
& One task with models varied;
  identically ordered images and independent SAEs
& \makecell[tl]{L1--L3 (primary)\\L4 \\(supplementary)}
& dCor (Eq.~\eqref{eq:dcor}),
  NNGS (Eq.~\eqref{eq:nngs}),
  PCA+PAR (Eq.~\eqref{eq:procrustes}),
  TSI (Eq.~\eqref{eq:tsi}) and feature--task MI (Eq.~\eqref{eq:mi-feature-task}),
  RBF-MMD (Eq.~\eqref{eq:mmd}),
  \newline $r$-Wasserstein distance (Eq.~\eqref{eq:wasserstein}),
  bottleneck distance (Eq.~\eqref{eq:bottleneck}) \\

Different Tasks, Same Model
& One model with tasks varied
& L1 (primary)
& Inter-task cosine (Eq.~\eqref{eq:inter-task-pair-cosine}),
  MDS (visualization) \\

Different Tasks, Different Models
& Tasks and models varied; no image-ID or SAE-coordinate alignment
& \makecell[tl]{L3 (primary)\\L4 (supplementary)}
& Task-conditioned activation-rate MMD extension of Eq.~\eqref{eq:mmd},
  averaged reciprocally and then ranked within each model pair;
  \newline $H_0/H_1$ $r$-Wasserstein and bottleneck distances
  (Eqs.~\eqref{eq:wasserstein} and~\eqref{eq:bottleneck}) \\

\bottomrule
\end{tabular}
\vspace{-12pt}
\end{table*}

\subsection{Visualization Results and Illustrations}
\label{app:vis}

\makeatletter
\setlength{\@fptop}{0pt}
\setlength{\@dblfptop}{0pt}
\makeatother

This section collects the supplementary visual evidence referenced by the main results. The figures examine task-wise local dimensionality, its relationship to cross-model alignment, and within-model inter-task organization, and report the Level-3 SAE and Level-4 topological analyses for Conditions II and IV\@. Where applicable, captions state the unit of analysis, aggregation rule, and uncertainty summary needed to interpret the corresponding panel.

\subsubsection{Condition I: Same Task, Same Model}
Condition I holds the task and the model fixed and varies only the input image, asking whether the representation an instruction induces is a stable property of the task or is dominated by image content. It evaluates Level~1 through the intra-task similarity in Eq.~\eqref{eq:intra-task-pair-cosine} and Level~2 through the LID in Eqs.~\eqref{eq:lid}--\eqref{eq:mean_lid}. Both figures analyze the instruction-induced displacement, defined as the difference between the last-token hidden states obtained under the task prompt versus the neutral prompt (for Qwen-Image-Edit, the difference of spatially averaged condition-image features), computed over $b = 10$ frozen sampling repeats of $n = 100$ distinct images per task.

Figure~\ref{fig:figure2} evaluates Eq.~\eqref{eq:intra-task-pair-cosine} at every layer, plotting one curve per task against relative depth. The results indicate that within-task consistency varies across layers. Specifically, in five models the task curves begin at $0.95$ or above at the first layer and separate only with increasing depth, whereas Qwen-Image-Edit starts lower for some tasks and fluctuates more across depth. Emu3.5 is the exception: it exhibits a sharp early rise, before showing greater separation among task curves at later layers. Consequently, the $0.80$--$0.96$ interval quoted at the analysis layer does not describe the network as a whole.

\begin{figure*}[t]
    \centering
    \includegraphics[width=\textwidth]{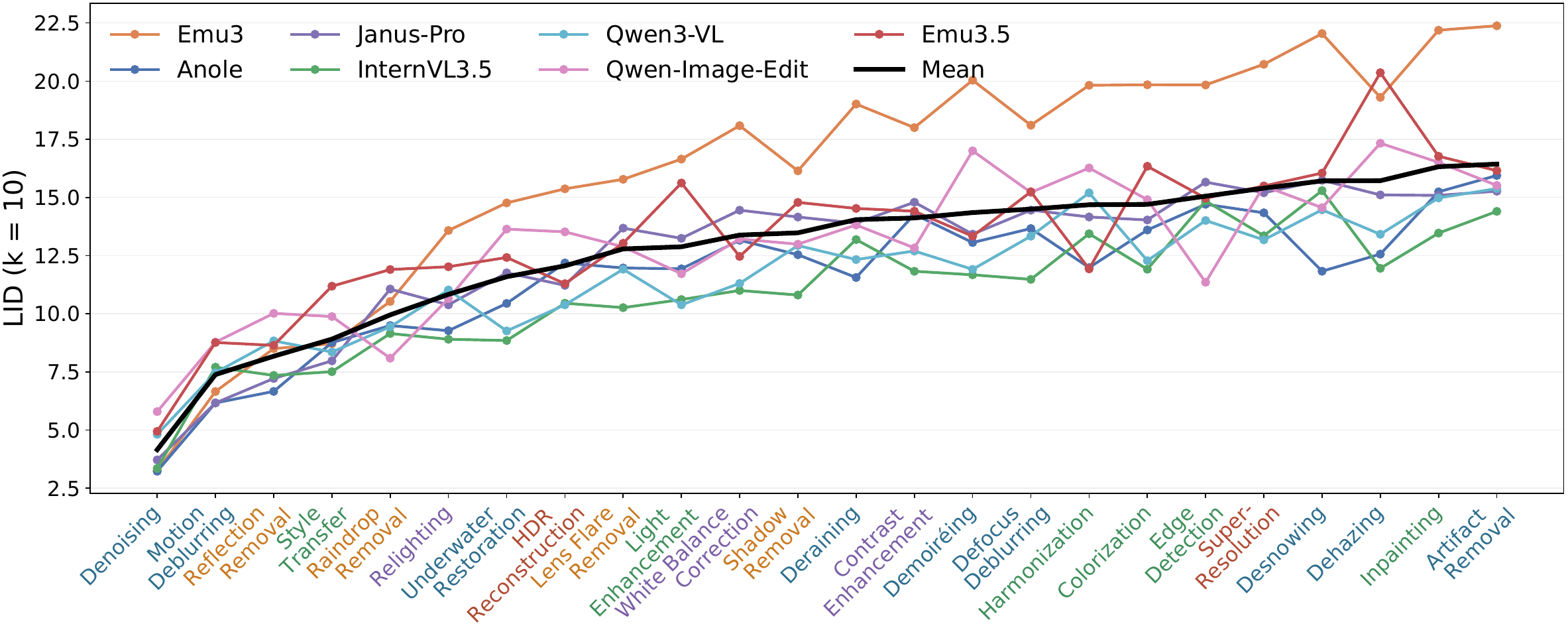}
    \caption{\textbf{Condition I: Same Task, Same Model.} LID per task, estimated with the Levina--Bickel estimator at $k = 10$ at each model's analysis layer $\ell^*$. Tasks are ordered by their cross-model mean, shown as the black line.
    }
    \label{fig:figure1_Appendix}
    \vspace{-8pt}
\end{figure*}

Figure~\ref{fig:figure1_Appendix} evaluates Eqs.~\eqref{eq:lid}--\eqref{eq:mean_lid} on these same displacements at the single layer $\ell^*$ chosen by Eq.~\eqref{eq:frobenius-evolution}, applying the Levina--Bickel estimator at $k = 10$. Furthermore, each sampling repeat contains no duplicate images, as the estimator degenerates when the nearest-neighbor distance vanishes. Tasks are ordered by their cross-model mean, yielding an ordering that remains nearly identical across models (mean pairwise Spearman $\rho = 0.80$, with Denoising lowest in all seven). This indicates that local complexity is governed by the task rather than the architecture, although the estimator is downward-biased with $n = 100$ samples in a space of several thousand dimensions and only the ordering is interpreted.

\begin{figure*}[t]
    \centering
    \includegraphics[width=\textwidth]{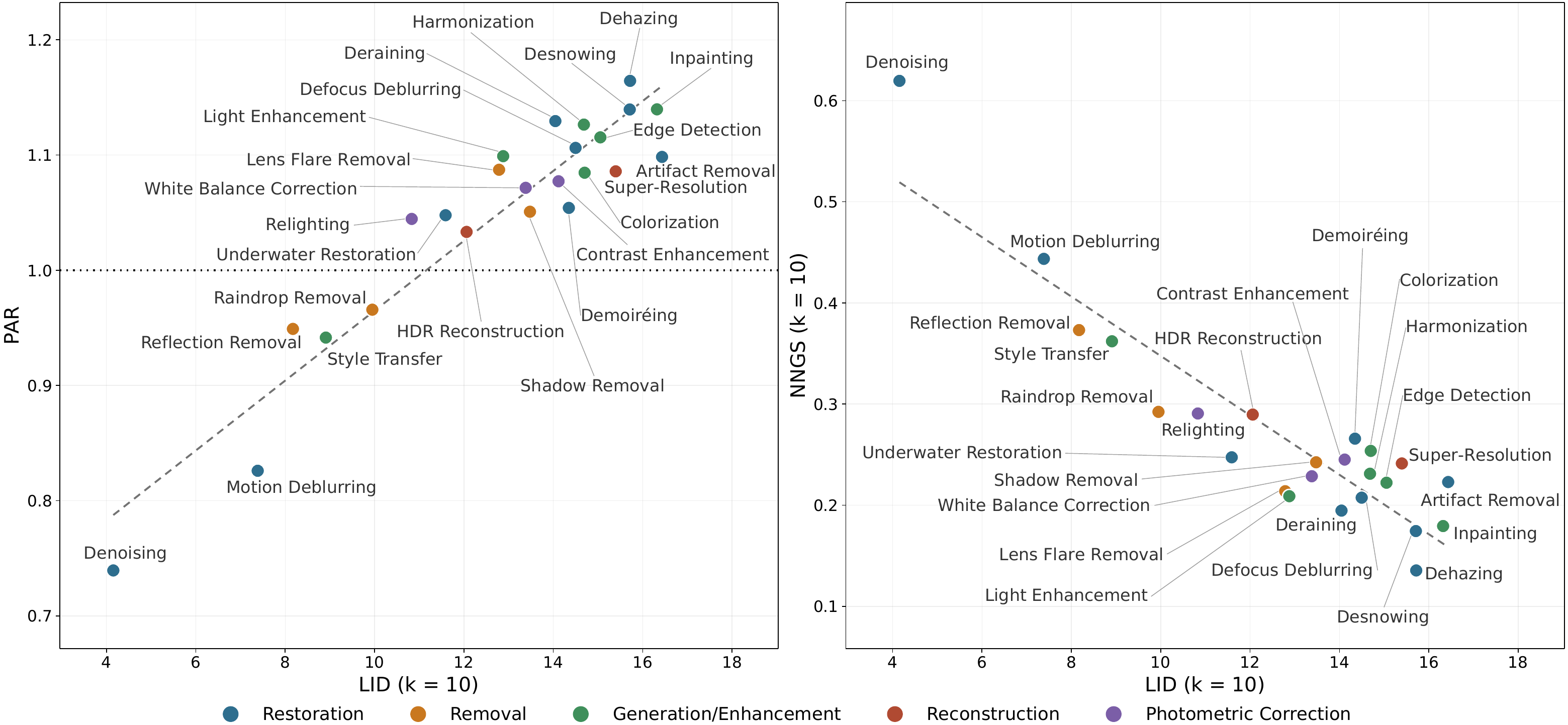}
    \caption{\textbf{Condition II: Same Task, Different Models.} Local dimensionality against cross-model agreement, one point per task. The abscissa is the seven-model mean LID of Figure~\ref{fig:figure1_Appendix}. The ordinates are the held-out PAR (left) and the chance-adjusted neighborhood overlap (right), each averaged over the 21 model pairs. The dotted line marks the PAR baseline of 1.0, the residual obtained by predicting the mean.
    }
    \label{fig:figure2_1_Appendix}
    \vspace{-10pt}
\end{figure*}

\begin{figure*}[ht]
    \centering
    \includegraphics[width=\textwidth]{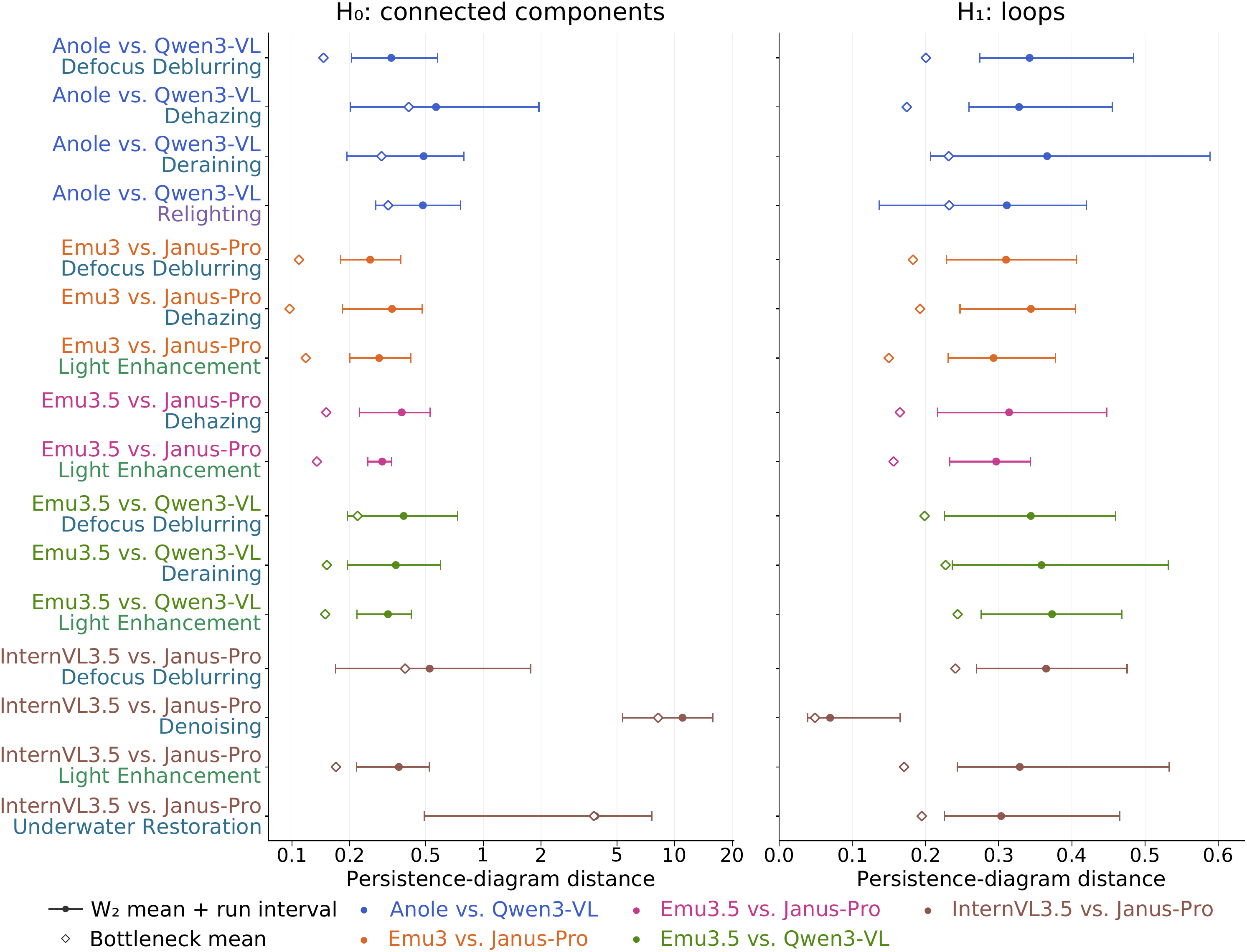}
    \caption{\textbf{Condition II: Same Task, Different Models.} Level-4 2-Wasserstein and bottleneck distances for $H_0$ and $H_1$ across 16 model-pair and task combinations.}
    \label{fig:figure2_2_Appendix}
    \vspace{-14pt}
\end{figure*}

\subsubsection{Condition II: Same Task, Different Models}
Condition II holds the task fixed while varying the model, testing whether different architectures arrive at similar geometry for the same task. The sampling manifests ensure that the $i$-th row of every representation matrix across models corresponds to the same image, providing the pairing required by dCor, NNGS, and PAR.
Figure~\ref{fig:figure3} reports four metrics over the 21 model pairs, averaged across the 24 tasks and ordered by dCor.
PAR is computed after projecting each model's representations separately onto its top 64 principal components, and Level~4 uses the topological pipeline in Section~\ref{sec:level4}.

Agreement turns out to differ between global and local comparisons. The dCor in Eq.~\eqref{eq:dcor} averages 0.77, whereas the chance-adjusted NNGS averages only 0.27. We use $k=10$ for both LID and NNGS and report $(\mathrm{NNGS}-c)/(1-c)$, where $\mathrm{NNGS}$ is the raw overlap defined in Eq.~\eqref{eq:nngs} and $c=k/(n-1)\approx0.101$ for $n=100$.
The PAR in Eq.~\eqref{eq:procrustes} is likewise reported in a held-out form: the orthogonal map is fitted on 80\% of the rows and scored on the remaining 20\%, with the test residual normalized by the Frobenius norm of the transformed held-out target matrix. Predicting the training-set target mean gives zero in the centered PCA space and therefore a normalized residual of 1.0. The residual averages 1.049, with 16 of the 21 pairs scoring above that 1.0 line. This indicates that an orthogonal alignment estimated on part of the data fails to generalize to the rest. The models therefore share a coarse ordering of distances without sharing either local neighborhoods or a coordinate frame.

Figure~\ref{fig:figure2_1_Appendix} isolates the one systematic exception by plotting each task's seven-model mean LID against its cross-model agreement. This yields a correlation of $\rho = +0.853$ against PAR and $\rho = -0.764$ against neighborhood overlap, with Denoising and Motion Deblurring the only tasks whose residuals fall markedly below 1.0 (at 0.74 and 0.83), whereas every task with an LID exceeding 15 scores above 1.1. This suggests that a task with fewer operative directions leaves the models less room to differ.

Figure~\ref{fig:figure4} shows the Level~3 results. For each of the 21 model pairs, we compute the same-task cross-model activation-rate MMD for all 24 tasks and average the results across tasks. We then compare this value with a matched within-model baseline computed from different-task pairs, averaging results over three SAE random seeds. Under this comparison, 6 model pairs satisfy the criterion after seed aggregation, while 3 satisfy it consistently across all three seeds. The ratios in the left panel are the same-task cross-model MMD divided by the matched within-model, different-task reference; only ratios below one are labeled. The TSI and feature--task MI panels instead characterize task selectivity within each individual model without requiring SAE features to be aligned across models.

For Level~4, $H_0$ and $H_1$ denote the two homology degrees reported throughout this section. As the filtration radius increases, $H_0$ records connected components and their merging, summarizing multiscale cluster and connectivity structure, while $H_1$ records the birth and death of one-dimensional loops. We report $H_0$ and $H_1$ separately rather than averaging them. A high $H_0$ distance with a low $H_1$ distance, for example, indicates different component structure but similar loop structure.
Figure~\ref{fig:figure2_2_Appendix} presents the supplementary Level-4 analysis for 16 evaluated model-pair and task combinations. For both $H_0$ and $H_1$, it reports mean 2-Wasserstein and bottleneck distances over $b=10$ frozen sampling runs, each with $n=100$ samples per task. Horizontal error bars show run-to-run variation in the 2-Wasserstein distance, and endpoints are the 2.5th and 97.5th percentiles of the ten run-level distances.
The 2-Wasserstein distance agrees in direction while supplying a complementary topological check.

\begin{figure*}[t]
    \centering
    \includegraphics[width=\textwidth]{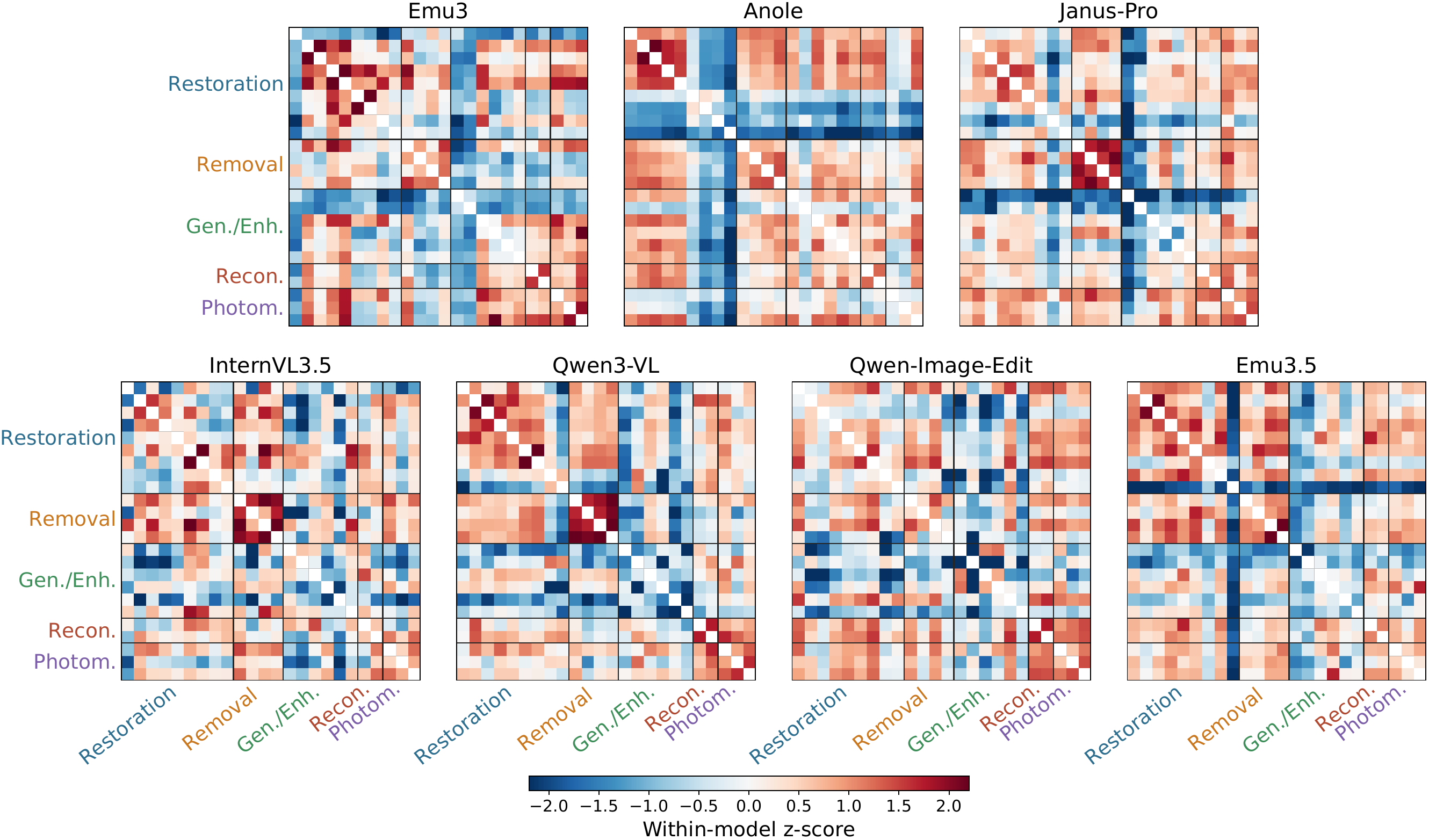}
    \caption{\textbf{Condition III: Different Tasks, Same Model.} Inter-task cosine similarity matrices, with one panel per model. Both axes carry the same 24 tasks in a fixed order grouped by perceptual family, with rules and labels marking the five categories. Each panel is standardized on its own off-diagonal entries, so color encodes how far a task pair departs from the typical pair of that model, red above and blue below, on the common scale of $\pm 2.2$ SD given in the color bar. The diagonal is left blank because it carries intra-task rather than inter-task similarity.
    }
    \label{fig:figure3_Appendix}
    \vspace{-10pt}
\end{figure*}

\subsubsection{Condition III: Different Tasks, Same Model}
Condition III holds the model fixed while varying the task, asking how the 24 tasks are organized relative to one another inside a single network. It draws on Level~1 through the inter-task similarity of Eq.~\eqref{eq:inter-task-pair-cosine}, whose collection over all task pairs forms the layer-wise task similarity matrix.

Figure~\ref{fig:figure3_Appendix} shows that matrix at the analysis layer for every model.
Because the raw ranges differ widely between models, such as 0.29--0.65 in Emu3 against 0.52--0.86 in Qwen3-VL, each panel is standardized on its own off-diagonal entries before plotting. What is compared across panels is therefore block structure rather than overall level. Under this standardization, the Removal block remains coherent in every model, whereas Generation/Enhancement shows no consistent block.

Figure~\ref{fig:figure5} aggregates the seven standardized matrices into a consensus, taking their mean across models as the signal and their spread as the noise. It draws an edge only where the consensus exceeds 0.714 SD, which represents the mean disagreement across models, ensuring the average signal is at least as large as the typical cross-model dispersion. We find 36 of the 276 task pairs qualify, and 20 of them cross a family boundary.
The family gap is the mean consensus for within-family pairs minus that for cross-family pairs.

The strongest are Defocus--Motion Deblurring at $+1.66$, Deraining--Desnowing at $+1.63$, and Reflection--Shadow Removal at $+1.61$, representing pairs that share a degradation mechanism.
The cross-family edges show that relations in the graph are not confined to family boundaries. Seven tasks carry no edge for different reasons. Colorization and Harmonization sit below their model average similarity to every other task, with best consensus at $-0.29$ and $-0.26$, whereas Inpainting misses the threshold by 0.009 and is better read as undecided than as isolated.

Unlike t-SNE~\citep{vandermaaten2008visualizing} or UMAP~\citep{mcinnes2018umap}, classical MDS is deterministic and embeds the dissimilarities directly, which matters here because the relation graph contains only 24 nodes, a regime in which a t-SNE or UMAP layout is not stable.
Node positions derive from MDS on the consensus matrix, whose first two coordinates carry 31.5\% of the structure with a rank correlation of 0.772 between plotted and true dissimilarity. Therefore, the layout should be read for grouping rather than for precise metric distance.
Positions and edges encode different information, since coordinates summarize a task's profile against all 23 others while an edge reflects a single pairwise value.

\subsubsection{Condition IV: Different Tasks, Different Models}
\begin{figure*}[hb]
    \centering
    \includegraphics[width=\textwidth]{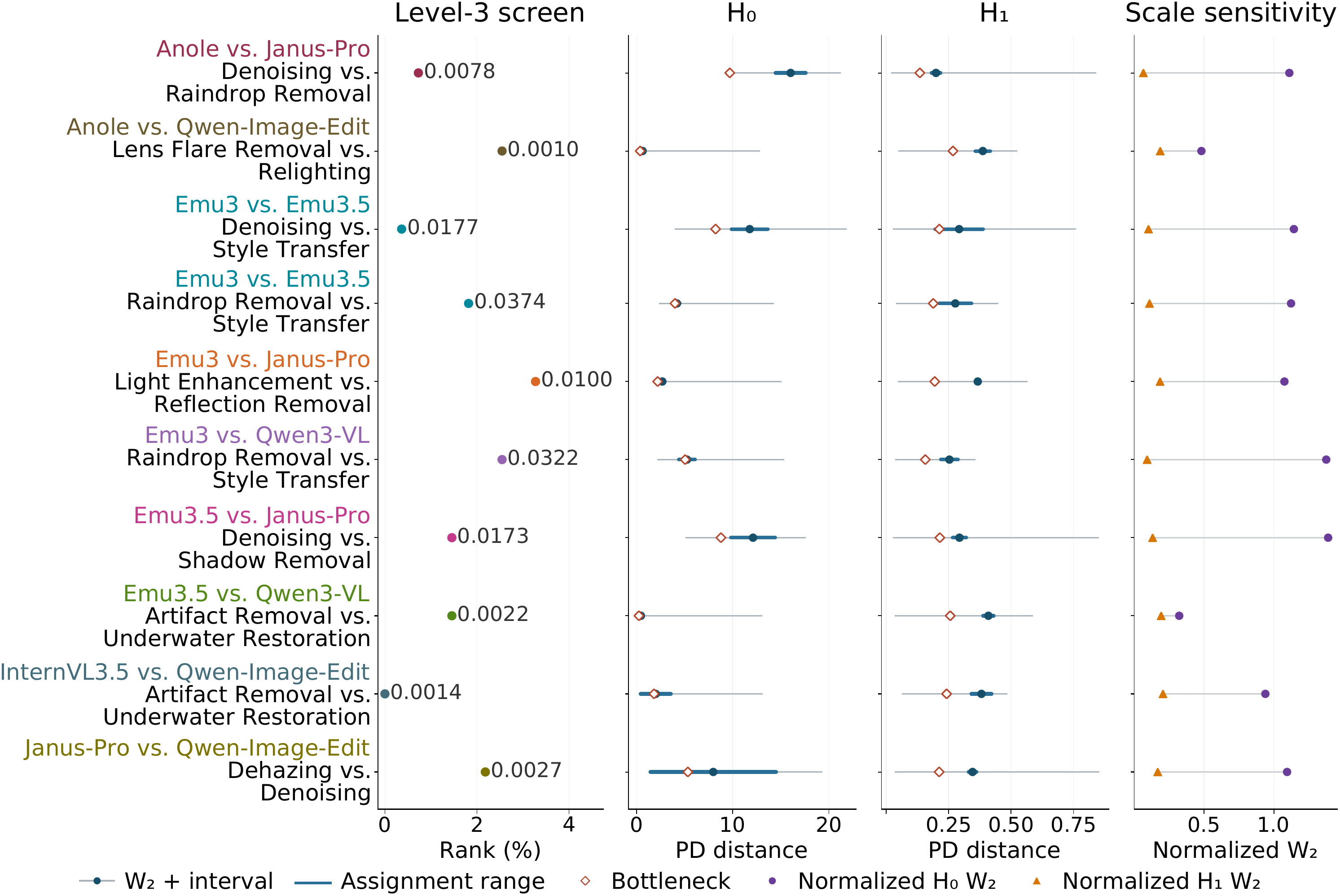}
    \caption{\textbf{Condition IV: Different Tasks, Different Models.} Level-3 screening rank and Level-4 $H_0/H_1$ topological distances for the 10 reciprocal candidates, shown on raw and scale-normalized views. Dots show 2-Wasserstein means over the ten sampling runs and both task assignments, thin bars span the bootstrap percentile intervals of both assignments, and thick bars show the range between their means.}
    \label{fig:figure4_Appendix}
    \vspace{-10pt}
\end{figure*}

Condition IV changes both task and model without assuming paired images or aligned SAE coordinates.
Figure~\ref{fig:figure4_Appendix} follows the 10 reciprocal candidates retained by the exploratory Level-3 screen.
The first panel gives their within-model-pair percentile ranks; its numerical annotations are raw reciprocal MMD values. The $H_0$ and $H_1$ panels report 2-Wasserstein means and bootstrap percentile intervals, the range across the two task assignments, and bottleneck distances. The final panel shows scale-normalized distances: the pairwise-distance matrix of each point cloud is divided by its own median nonzero distance before computing persistent homology.

Topology uses $n=100$, $b=10$, and $B_{\mathrm{PH}}=50$, remaining a supplementary and unpaired comparison. For each assignment, interval bounds represent the 2.5th and 97.5th percentiles of the 50 bootstrap distances. The displayed interval spans from the lower of the two lower bounds to the higher of the two upper bounds.

\end{document}